\pdfoutput=1

\documentclass[11pt]{article}

\usepackage[hyperref,acceptedWithA]{tacl2021v1}

\usepackage{times}
\usepackage{latexsym}
\usepackage{comment}
\usepackage{subcaption}
\usepackage[font=small,labelfont=bf]{caption}

\usepackage[T1]{fontenc}

\usepackage[utf8]{inputenc}

\usepackage{microtype}

\usepackage{inconsolata}
\usepackage{todonotes}

\usepackage{xspace}
\usepackage{amsmath}
\usepackage{amssymb}
\usepackage{amsthm}
\usepackage{amsfonts}
\usepackage{tikz}
\usepackage{bbm}
\usepackage{xspace}
\usepackage{xcolor}
\usepackage{xfrac}

\usepackage[normalem]{ulem}
\usepackage{thmtools}
\usepackage{thm-restate}
\usepackage{enumitem}
\usepackage{booktabs}
\usepackage{multirow}
\usepackage{graphicx}
\usepackage{float}
\usepackage{listings}

\usepackage{stfloats}
\usepackage{ragged2e}

\usepackage{pifont}
\usepackage{cleveref}
\crefname{section}{\S}{\S\S}
\Crefname{section}{\S}{\S\S}
\crefname{table}{Tab.}{}
\crefname{figure}{Fig.}{Figs.}
\crefname{algorithm}{Alg.}{}
\crefname{equation}{Eq.}{}
\crefname{appendix}{App.}{}
\crefname{thm}{Theorem}{}
\crefname{hypot}{Hypothesis}{}
\crefname{question}{RQ}{}
\crefname{prop}{Proposition}{}
\crefname{cor}{Corollary}{}
\crefname{observation}{Observation}{}
\crefname{assumption}{Assumption}{}
\crefformat{section}{\S#2#1#3}
\crefformat{footnote}{#2\footnotemark[#1]#3}

\newtheorem{question}{RQ}

\def\special#1{\textsc{#1}}

\newcommand{\citeay}[1]{\citeauthor{#1}, \citeyear{#1}}
\newcommand{\newterm}[1]{\textbf{#1}}

\newcommand{\lone}{\ensuremath{\mathrm{L}_1}\xspace}
\newcommand{\ltwo}{\ensuremath{\mathrm{L}_2}\xspace}
\newcommand{\epochs}{\ensuremath{\mathrm{E}}\xspace}
\newcommand{\size}{\ensuremath{\mathrm{S}}\xspace}

\newcommand{\gpt}{GPT-2\xspace}
\newcommand{\roberta}{RoBERTa\xspace}
\newcommand{\StrongInnate}{Strong Innate Claim\xspace}
\newcommand{\strongInnate}{strong innate claim\xspace}
\newcommand{\StrongExperience}{Strong Experiential Claim\xspace}
\newcommand{\strongExperience}{strong experiential claim\xspace}

\newcommand{\cp}{CP\xspace}

\newcommand{\bpe}{BPE\xspace}
\newcommand{\ewc}{EWC\xspace}
\newcommand{\lm}{LM}

\newcommand{\ppl}{PPL\xspace}
\newcommand{\blimp}{BLiMP\xspace}
\newcommand{\glue}{GLUE\xspace}
\newcommand{\sglue}{SuperGLUE\xspace}

\newcommand{\learningalg}{\ensuremath{\mathbb{L}}\xspace}

\DeclareMathOperator*{\E}{\mathbb{E}}

\newcommand{\appropto}{\mathrel{\vcenter{
  \offinterlineskip\halign{\hfil$##$\cr
    \propto\cr\noalign{\kern2pt}\sim\cr\noalign{\kern-2pt}}}}}

\newcommand{\colorprior}[1]{{ #1}}

\newcommand{\prior}{\colorprior{\pi}}

\newcommand{\strone}{\boldsymbol{x}}
\newcommand{\strtwo}{\boldsymbol{y}}
\newcommand{\parameters}{\Theta}
\newcommand{\regularizer}{\mathcal{R}}

\newcommand{\defeq}[0]{\mathrel{\stackrel{\textnormal{\tiny def}}{=}}}

\newcommand{\bigO}{\mathcal{O}}

\newcommand{\citeposs}[1]{\citeauthor{#1}'s (\citeyear{#1})}

\newcommand{\cutforspace}[1]{}

\newcommand{\vtheta}{\boldsymbol{\theta}}

\DeclareMathOperator*{\argmax}{\mathrm{argmax}}
\newcommand{\R}{\mathbb{R}}
\newcommand{\dataset}{\mathcal{D}}
\newcommand{\dataone}{\dataset_{\lone}}
\newcommand{\datatwo}{\dataset_{\ltwo}}

\newcommand{\fisher}{\mathbf{F}}

\newcommand{\thetaone}{\vtheta_{\lone}^{*}}

\def\gL{{\mathcal{L}}}

\newcommand{\en}{\mathtt{en}}

\newcommand{\de}{\mathtt{de}}
\newcommand{\nl}{\mathtt{nl}}
\newcommand{\es}{\mathtt{es}}
\newcommand{\pl}{\mathtt{pl}}
\newcommand{\el}{\mathtt{el}}
\newcommand{\ru}{\mathtt{ru}}
\newcommand{\fin}{\mathtt{fi}}
\newcommand{\tr}{\mathtt{tr}}
\newcommand{\ar}{\mathtt{ar}}
\newcommand{\ko}{\mathtt{ko}}
\newcommand{\java}{\mathtt{java}}

\newcommand{\gutenberg}{\ensuremath{\textsl{Gutenberg}}\xspace}
\newcommand{\subtitles}{\ensuremath{\textsl{OpenSubtitles}}\xspace}
\newcommand{\wikipedia}{\ensuremath{\textsl{Wikipedia}}\xspace}
\newcommand{\thestack}{\ensuremath{\textsl{The Stack}}\xspace}

\definecolor{codegreen}{rgb}{0,0.6,0}
\definecolor{codegray}{rgb}{0.5,0.5,0.5}
\definecolor{codepurple}{rgb}{0.58,0,0.82}
\definecolor{backcolour}{rgb}{0.95,0.95,0.92}

\lstdefinestyle{customstyle}{
    backgroundcolor=\color{white},   
    commentstyle=\color{codegreen},
    keywordstyle=\color{magenta},
    numberstyle=\tiny\color{codegray},
    stringstyle=\color{codepurple},
    basicstyle=\ttfamily\footnotesize,
    breakatwhitespace=false,         
    breaklines=true,
    postbreak=\mbox{\textcolor{red}{$\hookrightarrow$}\space},
    captionpos=b,                    
    keepspaces=true,                 
    numbers=left,                    
    numbersep=5pt,                  
    showspaces=false,                
    showstringspaces=false,
    showtabs=false,                  
    tabsize=2,
    frame=lines
}

\title{Investigating Critical Period Effects in Language Acquisition through Neural Language Models}

\author{
Ionut Constantinescu \quad  Tiago Pimentel \quad Ryan Cotterell \quad Alex Warstadt\\
ETH Z{\"u}rich \\
\url{ionut.constantinescu@alumni.ethz.ch} \quad \url{tiago.pimentel@inf.ethz.ch} \quad \\
\url{ryan.cotterell@inf.ethz.ch} \quad \url{awarstadt@ethz.ch}
}

\begin{document}
\maketitle

\begin{abstract}
Humans appear to have a critical period (CP) for language acquisition: 
Second language (\ltwo) acquisition becomes harder after early childhood, and ceasing exposure to a first language (\lone) after this period (but not before) typically does not lead to substantial loss of \lone proficiency.
It is unknown whether these \cp effects result from innately determined brain maturation or as a stabilization of neural connections naturally induced by experience.
In this study, we use language models (\lm s) to test the extent to which these phenomena are peculiar to humans, or shared by a broader class of language learners. 
We vary the age of exposure by training LMs on language pairs in various experimental conditions, and find that \lm s, which lack any direct analog to innate maturational stages, do not show \cp effects when the age of exposure of \ltwo is delayed.
Our results contradict the claim that \cp effects are an inevitable result of statistical learning, and they are consistent with an innate mechanism for \cp effects.
We show that we can reverse-engineer the \cp by introducing a regularizer partway through training to simulate a maturational decrease in plasticity. 
All in all, our results suggest that \lone learning on its own may not be enough to induce a \cp, and additional engineering is necessary to make language models more cognitively plausible.\looseness=-1
\end{abstract}

\section{Introduction}

The tension between nature and nurture is central to questions surrounding how humans acquire language.
The \newterm{Critical Period (\cp)} for language acquisition is no exception.
Around the onset of adolescence, humans exhibit a loss in ability to acquire a second language through immersion and a tendency not to forget their first language under deprivation \citep{penfield1959speech,lenneberg1967biological,johnson1989critical,pallier2003brain}.
Scholars of human development have long debated whether these phenomena are predetermined by innately encoded developmental changes in the maturing brain \citep{penfield1965conditioning,chomsky1965aspects,pinker1994language}, or natural consequences of increased experience that any typical statistical learner would be subject to \citep{elman1996rethinking,seidenberg2006connectionist,thiessen2016statistical}.\looseness=-1

Until recently, it was difficult to differentiate these two hypotheses; as we could only observe one kind of statistical learner (i.e., humans),
we could not identify which properties of its learning process were responsible for its behavior.
Recent improvement in neural language modeling upends this state of affairs \citep{warstadt2022what}.
\newterm{Language Models (\lm s)} can learn to simulate many native-like grammatical judgments \citep{warstadt-etal-2020-blimp-benchmark,zhang2021when,hu2020systematic}---long regarded as one of the main behavioral measures of native speaker knowledge \citep{chomsky1957syntactic,johnson1989critical}---and, like humans, they acquire this knowledge from unstructured input without the need for negative evidence. 
However, their learning algorithm and structure differ from those of humans in a number of ways.
\lm s can thus provide additional information about which phenomena are likely to be typical of general language learners, and which are peculiar to humans.\looseness=-1

\begin{figure*}[t]
    \centering
\includegraphics[width=.85\textwidth]{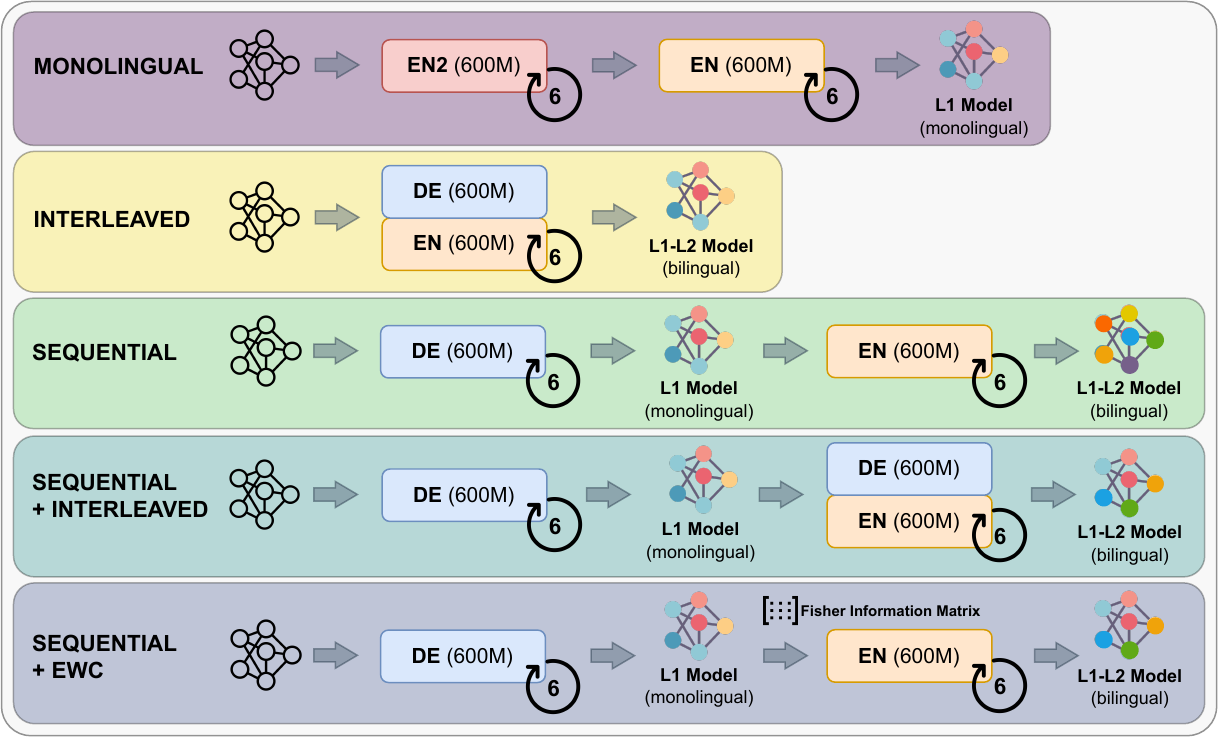}
    \caption{A visualization of the training conditions, using $\lone=\de$, $\ltwo=\en$, $\size=600$M, $\epochs=6$.}
    \label{fig:training-conditions}
    \vspace{-15pt}
\end{figure*}

In this work, we use language models to study the \cp for language acquisition, focusing on second language acquisition and first language attrition.
Our experiments test for \newterm{\cp effects}\footnote{The term \newterm{critical period} is sometimes used to refer to a specific biological construct. For clarity, we use the term \newterm{\cp effects} to refer to the characteristic observable effects of age of exposure on \lone and \ltwo performance.} in \lm s by training them from scratch on bilingual data, varying only the age of exposure to \ltwo.
We test whether, like humans, \lm s learn \ltwo more easily when exposed to it simultaneously with \lone from the beginning of training, rather than when exposed to it only after learning \lone. Similarly, we test whether they fail to forget \lone after extensive training on it.
Experimentally, we find that \lm s are unlike humans in \emph{both} respects.\footnote{
We note that our results thus align with the well-known fact that language models: (i) are prone to catastrophic forgetting; and (ii) are good at transfer learning. Our experiments, though, support the novel conclusion that transfer performance (training in \lone followed by \ltwo) leads to similar or better results than jointly training on both languages.
}
Thus, our results contradict the view that \cp effects are an expected consequence of statistical learning, and they are consistent with (but only provide weak evidence for) the view that the \cp in humans is a biologically programmed developmental stage.\looseness=-1

The benefits of studying the \cp in neural models, however, go beyond just discerning the two hypotheses above.
If \lm s do differ from humans, it may be useful to attempt to reverse-engineer those learning properties exhibited by humans \citep{dupoux2018cognitive}.
Furthermore, minimizing differences between \lm s and humans is a necessary step in enabling their use more broadly as models of human language acquisition \citep{warstadt2022what}.
Thus, we also attempt to reverse-engineer a \cp by simulating a loss of neural plasticity using \newterm{Elastic Weight Consolidation} \citep[\newterm{\ewc};][]{kirkpatrick2017overcoming}, a Bayesian regularizer used in machine learning to mitigate catastrophic forgetting.
Our experiments show that both of the \cp effects emerge in tandem when the model's plasticity is explicitly controlled in this way.
Our findings demonstrate the utility of \lm s as tools for theories about human language acquisition, and they suggest a path forward to making \lm s more developmentally plausible models of human language acquisition.\looseness=-1

\section{Background: The Critical Period}

The proposition that there is a critical period for language learning has long been prominent in language acquisition research \citep{penfield1959speech,lenneberg1967biological}.
Discussions around the \cp, however, typically cluster a number of related observations; these must be teased apart in order to be properly understood \citep{singleton2005critical,mayberry2018rethinking}.
The critical period can be divided into 3 main phenomena, namely: a \cp for \lone acquisition, a \cp for \ltwo acquisition, and a \cp for \lone attrition.
We focus on the latter two here.\footnote{
Strong evidence of a \cp for \lone acquisition has been demonstrated in late \lone learners from the deaf community \citep{mayberry1989looking,newport1990maturationala}.
However, as simulating late \lone exposure requires training on non-linguistic data, we consider it beyond the scope of this work.
}\looseness=-1

\subsection{Critical Period for \ltwo Acquisition}

\cp effects for \ltwo acquisition consist of greater difficulty in learning a second language and worse learning outcomes as the age of exposure increases.
As humans vary greatly in the beginning of \ltwo exposure, this is perhaps the most well-known of the \cp phenomena.
The effects of age of exposure on phonetics and phonology (i.e., one's accent) are part of folk knowledge and were a key piece of evidence in the first works to propose a neurological mechanism for these \citep{lenneberg1967biological}.
Numerous studies also show that age of exposure correlates with worse \ltwo performance on morphological and syntactic acceptability judgment tasks \citep{johnson1989critical,hartshorne2018critical}.
While the exact nature and reliability of these effects has been questioned at times \citep{ioup1994reexamining}, the existence of age-of-exposure effects is generally accepted. We refer the reader to several 
thorough reviews of the relevant evidence \citep{singleton2005critical,thiessen2016statistical,mayberry2018rethinking}.
\looseness=-1

In the realm of computational learners, no prior work has tested this \cp in a controlled manner.
In a more general form, however, \ltwo acquisition has been studied in depth \citep[\textit{inter alia}]{dufter-schutze-2020-identifying,chen2023improving}.
Most related to our work, \citet{oba2023second} trained a number of language models on an \lone and then fine-tuned these models on both \lone and \ltwo; they find that, unlike humans, this two-step training improves \lm s' \ltwo performance.
This already suggests that LMs may not show \cp effects for \ltwo learning.
However, they do make \ltwo learning relatively easier by fine-tuning their models on \lone and \ltwo simultaneously within the same bidirectional transformer context, which we contend is not cognitively plausible.\looseness=-1

Beyond \citeposs{oba2023second} study, weak evidence against the existence of a \cp for \ltwo in neural networks is suggested by a large body of work on transfer learning which fine-tunes pretrained neural networks to perform new tasks \citep[e.g.,][]{devlin-etal-2019-bert,roberta,driess2023palme}.
These studies indicate that a neural model can achieve superior performance on a fine-tuned task compared to training on it from scratch. However, these studies do not manipulate age-of-exposure while controlling for the total amount of input, and their main focus is on tasks other than language modeling itself, such as classification.

\subsection{Critical Period for \lone Attrition}

The \cp for \lone attrition refers to a loss of proficiency in \lone due to a lack of exposure to it.
This phenomenon is largely constrained to earlier ages \citep{pallier2007critical}, as adults who emigrate from their \lone community do not typically forget their \lone entirely. However, profound attrition is possible if \lone exposure ceases during childhood. For example, \citet{pallier2003brain} studied Korean-born adoptees in France who had no recognition of their \lone, despite living in Korea for as long as eight years.

In the computational domain, language attrition relates to another large body of work in life-long and continual learning.
In short, a large number of works have shown that neural networks are prone to catastrophic forgetting \citep{mccloskey1989catastrophic,french1999catastrophic}, losing most of their proficiency in their original training domain when fine-tuned on another.
Continual learning mitigates catastrophic forgetting through the use of adapters \citep{houlsby2019parameter,pfeiffer2020adapterhub}, regularizers \citep{kirkpatrick2017overcoming,pan2020continual}, or further training in the original domain.\looseness=-1

\subsection{Theories}

Critical period effects in humans are typically interpreted as evidence that neural plasticity in the language centers of the brain decreases as the brain matures \citep{newport1990maturationala}.
However, the cause of this decrease in plasticity is a matter of debate,
with much of the divergence among theories stemming from whether they emphasize \newterm{innate} or \newterm{experiential} mechanisms as responsible for this decrease.\footnote{With exceptions, most views are quite diverse and many scholars advocate for a nuanced view with multiple causes \citep[e.g.,~][]{newport1990maturationala,thiessen2016statistical,singleton2005critical}. 
Other explanations for \cp involve social factors, such as a decrease in willingness to experiment, 
in motivation to fit into one's community, or in the likelihood of being immersed in the target language \citep{hartshorne2018critical}.
}

Innate accounts of the \cp argue that this loss in plasticity is driven by properties which are specific to how humans acquire language.
Some of these accounts are based on the hypothesis that children---but not adults---are equipped with a specialized language acquisition device such as Universal Grammar \citep{chomsky1965aspects,newport1990maturationala}.  
On this view, the \cp occurs when Universal Grammar is (wholly or partially) lost, displaced, or dismantled as we age, which would explain why adults struggle with language acquisition (\citeay{chomsky1965aspects}, p.~207; \citeay{borer1987maturation}; \citeay{bley-vroman1988accessibility}; \citeay{schachter1988second}; \citeay{pinker1994language}, p.~294).\footnote{Child and adult language acquisition are seen as driven by different mechanisms on this view \citep{thiessen2016statistical}, supported by evidence that general analytic ability predicts adult---but not child---\ltwo learning outcomes \citep{dekeyser2000robustness}.\looseness=-1}
Other innate accounts are not language-specific, especially those with an explicit neurobiological basis.
For example, humans \citep[and other mammals;][]{paolicelli2011synaptic} go through a phase of synaptic pruning peaking in late childhood and adolescence \citep{huttenlocher1979synaptic,huttenlocher1990morphometric} during which disused neuronal connections are reduced \citep{hensch2005critical}.
Monolingual brains show signs of more extensive pruning than bilingual ones \citep{mechelli2004structural}, suggesting that early in life abundant synapses provide the necessary plasticity to acquire a second language with ease, and that later these synapses may be pruned if not yet recruited \citep{debot2006plastic}.
Beyond synaptic pruning, other neurobiological process such as myelination \citep{pulvermuller1994neurobiological,pujol2006myelination} and lateralization \citep{lenneberg1967biological} are also correlated with a loss in plasticity as we age.\footnote{Often, these processes have experiential correlates as well, i.e., their outcomes are modulated by experiences during development \citep{mechelli2004structural,cheng2019effects}.
This is consistent, however, with the timing and onset of the process being biologically determined.\looseness=-1
}\looseness=-1

By contrast, experiential accounts of the \cp argue that a loss in plasticity is a consequence of learning itself (\citeay{munro1986statedependent}; \citeay{elman1996rethinking}, p.~283; \citeay{ellis2000age}; \citeay{zevin2002age}; \citeay{seidenberg2006connectionist}; \citeay{thiessen2016statistical}; \citeay{achille2018critical}).
Early experiments on \newterm{connectionist} models found that stages associated with human development sometimes fall out naturally during the training of low-bias neural networks \citep{mcclelland1989parallel}.
Connectionist word learning simulations found an effect of age of acquisition on learning outcomes \citep{ellis2000age,zevin2002age}.
Numerous scholars explain this loss of plasticity---sometimes referred to as \newterm{entrenchment}---as a natural consequence of the training dynamics of networks that lead to convergence \citep{munro1986statedependent,elman1996rethinking,ellis2000age,seidenberg2006connectionist}.
As \citet[p.~1108]{ellis2000age} argue, in a model with random weights (e.g., after initialization) the activations of individual units tend towards intermediate values, leading to large weight changes, but as training proceeds, the units' activations tend towards extreme values making them less prone to change, even if the prediction loss is large.\looseness=-1

\section{The Role of LMs in Studying the \cp}\label{sec:role_of_lms_in_cp}

Computational models have the potential to be a powerful tool for informing debates about language acquisition, as they enable a degree of control over the learning mechanism and environment not possible with human subjects; their relevance to questions about human learning, however, is hampered by their numerous differences from human learners \citep{warstadt2022what}.
Nonetheless, there are some theoretical claims that current \lm s can provide strong or even conclusive evidence about.
Not surprisingly, these models are increasingly being used to test theories of language acquisition \citep{mccoy2020does,lavechin2021reverseengineering,wilcox2023using,warstadt2023findings}.
Language models can, for example, refute some poverty of the stimulus claims by providing existence proofs about language learnability \citep[p.~30]{clark2011linguistic}.\looseness=-1

In general, theories of \cp effects are rather diverse and nuanced.
However, we can identify two strong claims which are echoed in many of the accounts above and about which \lm s can provide evidence:
the \strongInnate and 
the \strongExperience.

\paragraph{\StrongInnate.}\!\!\textit{Innate learning constraints are necessary to explain critical period effects.}

The \strongInnate is implicit in the argument that the mere existence of \cp effects counts as evidence in favor of an innate mechanism like Universal Grammar \citep[see, e.g.,~][]{schachter1988second}.
This argument depends on the premise that a change in learning ability as extreme as what is seen in \ltwo acquisition could not (or would be very unlikely to) arise from a single domain-general learning mechanism.
This premise, and thus the argument, is simple to refute by finding a counterexample, that is, an instance of a low-bias learner that does show \cp effects.
Transformer-based \lm s, while not bias-free, have proven to be effective learners for vision \citep{dosovitskiy2021an}, protein folding \citep{jumper2021highly}, and many other types of data, suggesting that they are sufficiently domain-general to refute \strongInnate if they do show \cp effects.\looseness=-1

\vspace{-3pt}
\paragraph{\StrongExperience.}\!\!\textit{Critical period effects are a necessary consequence of successful statistical learning.}
\vspace{5pt}

The \strongExperience has been argued to follow from a mathematical understanding of the training dynamics of connectionist networks \citep{munro1986statedependent,ellis2000age}.
\citet{seidenberg2006connectionist} speak of a \newterm{paradox of success}, whereby successful generalization creates the conditions for a loss in plasticity.
This claim is similarly simple to refute, by finding a successful connectionist learner that fails to show \cp effects.

Studying the \cp in \lm s serves an additional purpose for our understanding of human language acquisition.
\citet{dupoux2018cognitive} argues that reverse-engineering properties of human language acquisition can give insights into the mechanisms behind those properties at the algorithmic or implementational level \citep{marr1976understanding}.
While we endorse this view, and do attempt to reverse-engineer \cp effects, our efforts are at the computational level.
The resulting models, however, do more closely resemble human learners in the relevant property, which makes results obtained from them more likely to generalize to humans \citep{warstadt2022what}.\looseness=-1

\section{Research Questions and Methodology}\label{sec:rqs}

To study the claims above, we now put forward two research questions which we investigate with the help of language models.

\begin{question}\label{hyp:cp_for_ltwo}
    Can we find evidence of a critical period for \ltwo learning in language models?
\end{question}

\begin{question} \label{hyp:cp_for_lone}
    Can we find evidence of a critical period for \lone attrition in language models?
\end{question}

These are the main questions that we want to investigate, and on which we focus our experiments.
We analyse them by first training \lm s in various multilingual setups---while altering the ages (epochs) at which \lone and \ltwo are acquired---and then evaluating our models on \lone and \ltwo.
Importantly, we do \emph{not} make any modifications to the \lm s' architecture or learning objectives in these experiments.\looseness=-1

Beyond these main questions, we also explore two other research question here.

\begin{question} \label{hyp:cp_with_ewc}
    Does reducing plasticity in language models induce human-like critical period effects?
\end{question}

\begin{question} \label{hyp:cp_language_pairs}
    Are critical period effects in language models dependent on \lone and \ltwo's similarity?
\end{question}

Investigating \cref{hyp:cp_with_ewc} serves a dual purpose.
First, it allows us to test whether the critical period effects associated with \ltwo acquisition and \lone attrition arise in tandem as a result of a loss in plasticity, or whether these phenomena can (at least in one case) be decoupled.
Second, if we are successful at reverse engineering these \cp effects, we obtain a model that more closely resembles human learners and that might be useful for future work.
We test this question by evaluating the linguistic performance on both \lone and \ltwo while training a model whose learning objective has an extra regularizer which enforces a reduction in plasticity.

Similarly, \cref{hyp:cp_language_pairs} also serves a dual purpose.
First, it informs us about the relationship between language similarity and \cp effects.
Second, it implicitly assesses how sensitive our results are to a specific choice of language pair, providing us with a notion of how robust our experiments are to this choice.
We test this question by performing the above analysis in a number of language pairs which differ in their similarity.

\subsection{Training conditions}\label{sec:training-conditions}

Our goal is to train a language model \learningalg from scratch on pairs of languages (\lone and \ltwo) while manipulating two independent variables: (i) the ages (epochs) of exposure to \lone and \ltwo, and (ii) the level of programmed plasticity. In this section our focus will be on providing the methodology for variable (i), whereas the methodology for variable (ii) is left to be presented in Section \ref{sec:ewc}.

The obvious way to manipulate age of acquisition in the case of language modeling is to alter the training data schedule. 
As visualized in \Cref{fig:training-conditions}, we consider five schedules, which we will refer to as ``training conditions'' throughout the paper.
Across all conditions, the datasets remain unchanged (for a given language pair) and the size of the training data per language is kept consistent (we denote this quantity by \size).
The total number of training iterations per example, also known as epochs, is also a constant number (we refer to this as \epochs). 
As a consequence, the amount of exposure to \lone and \ltwo is the same across conditions, with the inevitable exceptions of the \special{monolingual} and  \special{sequential-interleaved} conditions.\looseness=-1 

\paragraph{\special{monolingual}.}
This condition simulates a monolingual human learner exposed to only one language during their lifetime. 
The simplest approach in this condition would be to just train \learningalg for $2 \cdot \epochs$ epochs on a monolingual dataset. 
However, this would mean that \learningalg would be trained only on half the number of tokens ($\size$) compared to the other conditions ($2 \cdot \size$). 
To account for this, we create a second monolingual dataset of the same size and train \learningalg on the two datasets in a sequential manner.

\paragraph{\special{interleaved}.}
This condition aims to replicate a simultaneous bilingual human learner exposed to two different languages from birth. 
It is also an implementation of typical multitask learning.
We train \learningalg for a total of $\epochs$ epochs on an interleaved bilingual dataset. 
Throughout training, \learningalg encounters batches of fixed size that alternate between \lone and \ltwo. 
As the bilingual dataset is double in size compared to the monolingual datasets, $\epochs$ epochs provide the same amount of training steps per language as with the other conditions.\looseness=-1

\paragraph{\special{sequential}.}
This condition represents the experience of a late \ltwo learner who changes linguistic communities, losing exposure to \lone entirely.
It can also be seen as a typical implementation of transfer learning.
In this condition, \learningalg is trained for $\epochs$ epochs exclusively on \lone, and then subsequently trained for another $\epochs$ epochs on \ltwo. 
The shift from \lone to \ltwo occurs abruptly, with a complete halt in \lone exposure rather than a gradual transition.

\paragraph{\special{sequential-interleaved}.}
This condition is closely related to the previous sequential condition, but recreates an experience more common among human bilinguals where \lone exposure continues during \ltwo acquisition. 
It is also an implementation of one approach to continual learning.
After the initial stage of \lone learning, \learningalg is trained on \lone interleaved with \ltwo.
We continue to use the same \lone dataset from the initial training stage.

\paragraph{\special{sequential-ewc}.}
This condition attempts to emulate in \lm{s} an innate reduction in plasticity like that proposed for humans.
The \ltwo models are trained from the same \lone checkpoints as for the normal \special{sequential} condition, but with \ewc regularization added to the loss function. 
The  reduction in plasticity is not progressive, but only changes once, after \lone has been fully trained.

\subsection{Enforcing Plasticity} \label{sec:ewc}

There are several methods to simulate a computational reduction in plasticity in \lm s. 
We have chosen Elastic Weight Consolidation \citep{kirkpatrick2017overcoming} due to its popularity and simplicity.
\ewc introduces a Bayesian-inspired regularization term on a \lm{}'s loss partway through training; this term penalizes deviations from a prior distribution over the parameter space (defined in terms of \lone training), simulating the end of the critical period. The modified loss function is defined as
\begin{equation}\label{eq:ewc-loss}
   \gL(\vtheta) = \gL_{\ltwo}(\vtheta) + \lambda \cdot \regularizer_{\mathit{EWC}}(\vtheta) 
\end{equation}
We introduce an additional hyperparameter $\lambda \in \R_{\geq 0}$ to control the strength of the \ewc regularization term. In the trivial case when $\lambda=0$, there is no programmed decrease in plasticity. 
We provide the complete derivation of \ewc in \cref{app:ewc-derivation}.

\section{Experimental Setup\footnote{The code is released at \url{https://github.com/iconstantinescu/lm-critical-period}.}} \label{sec:experimental-setup}

This section provides a comprehensive description of the experimental setup for this project, including: languages, datasets, model architectures, and evaluation methods.

\paragraph{Languages.} 

Our experiments consist of training \lm s from scratch on language
pairs.
In this work, we rely on \texttt{English} ($\en$) data for evaluation, as it has a wealth of
well-studied resources for assessing language proficiency (see \Cref{sec:evaluation}).
Therefore, we justify the selection of the other languages in this study according to their relatedness to \texttt{English}. 
To reduce the computational overhead, we restrict most of our experiments to only two language pairs: \texttt{German}--\texttt{English} and \texttt{Finnish}--\texttt{English}. 
We choose these languages because they are both (relatively) high-resource languages that are well-represented in our chosen data domains. 
Furthermore, \texttt{German} ($\de$) is in the same language family as \texttt{English} (Indo-European/Germanic),  while \texttt{Finnish} ($\fin$) is unrelated to both (Finno-Ugric).

Nonetheless, to improve the generalizability of our results, we also run one experiment with an extended set of languages as \lone, using English as \ltwo (see Experiment 4 in \Cref{sec:experiment-4}).
We select languages from various language families (Indo-European (IE) or not) using different scripts (Latin (L) or not).
To represent the extreme endpoints, we also use a different corpus of English as the most closely related \lone, and a corpus in a programming language (\texttt{Java}) as the least closely related \lone.
The complete list is as follows (from most to least related):
\vspace{-5pt}
\begin{itemize}
    \itemsep-0.2em 
    \item Same language: \texttt{English2} ($\mathtt{en2}$)
    \item Germanic: \texttt{German} ($\de$), \texttt{Dutch} ($\nl$)
     \item IE, L: \texttt{Spanish} ($\es$), \texttt{Polish} ($\pl$)
     \item IE, Non-L: \texttt{Greek} ($\el$), \texttt{Russian} ($\ru$)
     \item Non-IE, L: \texttt{Finnish} ($\fin$), \texttt{Turkish} ($\tr$)
     \item Non-IE, Non-L: \texttt{Arabic} ($\ar$), \texttt{Korean} ($\ko$)
     \item Programming language: \texttt{Java} ($\java$)
\end{itemize}
\vspace{-5pt}

\paragraph{Datasets.}

Ideally, we would train on data that closely resembles the kinds of language children encounter during language acquisition. This typically involves natural speech and narratives. Unfortunately, there are no existing datasets in any language that are fully representative of the type and volume of language input a child is exposed to during learning, let alone in a variety of languages. As a consequence, in this work, we construct a customized mix of multilingual training data sourced from three complementary domains: \textbf{spoken}, \textbf{literature} and \textbf{non-fiction}.
We select the \subtitles \citep{lison2016opensubtitles2016} corpus for the spoken domain, the \gutenberg  \citep{gerlach2018standardized} collection for literature, and the \wikipedia content for non-fiction.\footnote{
To perform Experiment 4 with a broader range of languages, we exclude the \gutenberg dataset due to insufficient data availability across languages. 
Further, for the special \texttt{Java} (programming) language, we use an additional corpus called \thestack \citep{kocetkov2022stack}.}
While these choices of datasets are not fully developmentally plausible, we judge them to have a good balance of diversity, quality, and quantity among the limited set of publicly available multilingual datasets.\cutforspace{\footnote{Among these considerations, quantity requires some further discussion. While we know that English-speaking children will have been exposed to about 50M words before the end of the \cp \citep{gilkerson2017mapping}, previous work has found that \lm s are less data-efficient than humans \citep{zhang2021when,warstadt2022what}.
Therefore, if we want to maximize our chance of observing a human-like \cp effect in \lm s, we should aim to train models on no less than 50M words of data (per language), and preferably on much more, so that they can achieve native-like proficiency.}}
See \cref{app:data_preprocess} for preprocessing details and \cref{app:data} for data statistics.\looseness=-1

\paragraph{Models.} 
We run our experiments with both autoregressive and masked language models.
To represent these two categories, we consider \roberta \citep{roberta}, an encoder-only transformer trained to predict masked tokens, and \gpt \citep{radford2019language}, a decoder-only transformer trained to predict next tokens.
We rely on implementations from the HuggingFace Transformers library \citep{wolf-etal-2020-transformers}, namely \texttt{roberta-base} with 125M parameters and \texttt{gpt2} with 137M parameters.
We choose these two models due to their availability, size and scientific relevance.
These models are then trained according to the conditions defined in \cref{sec:training-conditions}.
In all three sequential conditions, \ltwo training starts from the final \lone checkpoints, but with a new optimizer.\looseness=-1

\paragraph{Hyper-parameters.}

We run a Bayesian hyperparameter search using the Weights \& Biases Sweeps API \citep{wandb} to identify good model configurations for our training data and methodology. 
We extract several model configurations (see \cref{app:hyperparams}) with which we run the various experiments from \cref{sec:experiments}.\looseness=-1

\paragraph{EWC Implementation.} \label{sec:ewc-implementation}

We estimate the Fisher Information Matrix according to \cref{eq:hessian-approx}, using $K=10$ samples per input.
The model is trained using the loss in \cref{eq:ewc-loss}.\footnote{While we use \ewc's loss for training, we still report the cross-entropy loss for evaluations to be consistent.}
We choose the regularization strength $\lambda$ such that \lone performance matches \ltwo performance at the end of training, which we identify experimentally to be $\lambda = 20$ for \gpt and $\lambda = 150$ for \roberta. We discuss the details and reasoning behind the choice of $\lambda$ in \cref{sec:experiment-2}.

\paragraph{Evaluation.}\label{sec:evaluation}

We use Perplexity (\ppl), \blimp and \glue to assess models' language proficiency throughout our experiments. To account for the differences in tokenization across language pairs, we report the \ppl per character.\footnote{%
In addition, \ppl per character being smaller in magnitude than the more familiar \ppl per token, differences between \ppl scores \emph{relative to the total magnitude} are smaller as well.
}
\blimp \citep{warstadt-etal-2020-blimp-benchmark} is a dataset of minimal pairs targeting contrasts in acceptability for a variety of grammatical phenomena in English.
Evaluation is performed in a zero-shot  setting by comparing LM surprisal on a grammatical and ungrammatical sentence.
\glue and superGLUE\footnote{We henceforth use \glue to refer to a combination of tasks from GLUE and superGLUE (see \cref{app:eval}) implemented in the BabyLM evaluation pipeline \citep{warstadt2023findings}.\looseness=-1} are compilations of semantic, commonsense, and syntactic tasks \citep{wang2018glue,wang2019superglue}.
Evaluation is performed by fine-tuning all parameters of the model as well as a randomly initialized classifier MLP.
While similar benchmarks (some even using the names \blimp and \glue) are available for many of the languages we study, these are inherently different datasets which are not mutually comparable.
Thus, throughout this work, we use only \texttt{English} as the evaluation language.
This is far from optimal from the perspective of linguistic representation, but we prioritize controlled evaluation, rather than obtaining a large set of exploratory results.\looseness=-1

\section{Experiments} \label{sec:experiments}

This section describes our four experiments, designed to provide evidence for answering the research questions detailed in \cref{sec:rqs}.

\subsection*{Experiment 1: Regular training. \\
$\mathbf{\lone \in \{\de, \fin\}}$, 
$\mathbf{\ltwo \in \{\en\}}$,
$\mathbf{\size = 600M}$,
$\mathbf{\epochs = 6}$.} \label{sec:experiment-1}

\begin{figure}[t!]
    \begin{subfigure}[]{\columnwidth}
        \centering
        \includegraphics[width=\columnwidth]
        {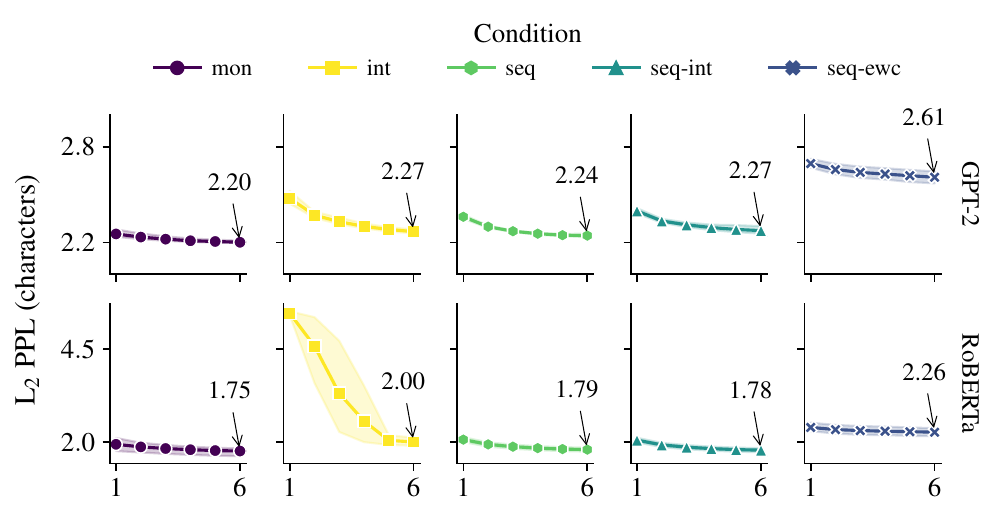}
        \label{fig:loss_l2_results}
    \end{subfigure}

    \vspace{-5pt}
     \begin{subfigure}[]{\columnwidth}
        \centering
        \includegraphics[width=\columnwidth]{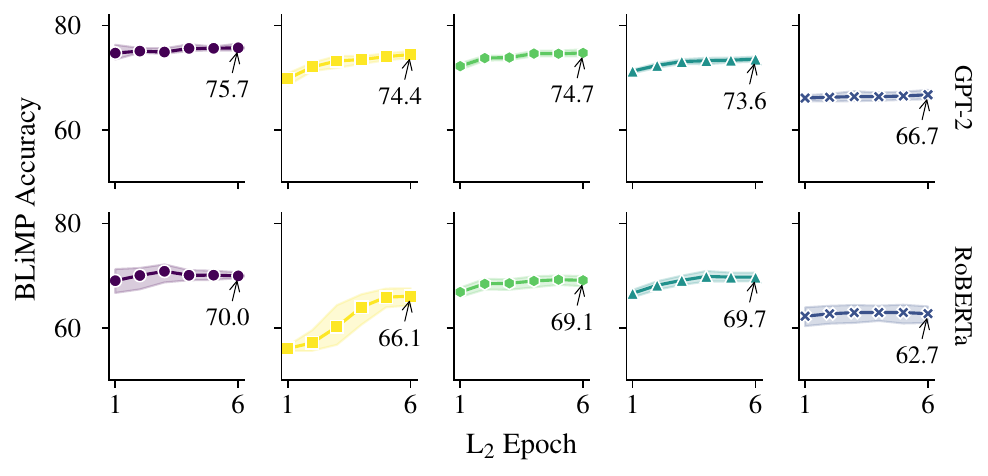}
        \label{fig:blimp_l2_results}
    \end{subfigure}
    
    \vspace{-10pt}
    \begin{subfigure}[]{\columnwidth}
        \centering
        \includegraphics[trim={0 0 0 0},clip,width=\columnwidth]{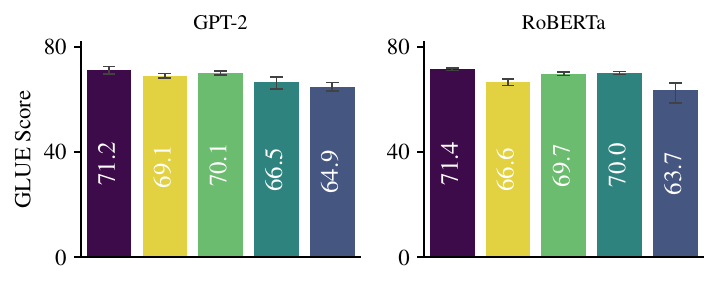}
        \label{fig:glue_l2_results}
    \end{subfigure}
    \vspace{-20pt}
    \caption{\ltwo ($\en$) results for regular training (6 epochs). Results are aggregated across model configuration and \lone ($\de$ and $\fin$). \textbf{Top:} \ppl per character on \ltwo ($\en$) during training on \ltwo. \textbf{Middle:} Accuracy on \blimp during training on \ltwo. \textbf{Bottom:} Performance on \glue at the end of training.\looseness=-1}
    \label{fig:l2-results}
  \vspace{-15pt}
\end{figure}

The goal of this setup is to study the \cp for \ltwo learning, which relates to \cref{hyp:cp_for_ltwo} and \cref{hyp:cp_with_ewc}. 
We run this experiment on both \gpt and \roberta models, using \texttt{German} and \texttt{Finnish} data as a first language and \texttt{English} data as a second language. 
The dataset for each language has the same size of 600 million tokens, a factor which was limited by the availability of the \texttt{Finnish} data. The models are trained with a limited budget of 6 epochs per language. This number has been empirically determined (after preliminary exploration) to provide a good trade-off between computational costs and model performance (i.e., the models perform sufficiently well after training for 6 epochs and the learning slows down). Finally, in order to introduce more variability and provide more result samples, the trainings are run with three different configurations ($\mathbf{C_1, C_2, C_3}$, see \cref{app:hyperparams}).

\begin{figure}[t!]
    \begin{subfigure}[]{\columnwidth}
        \centering
        \includegraphics[width=\columnwidth]{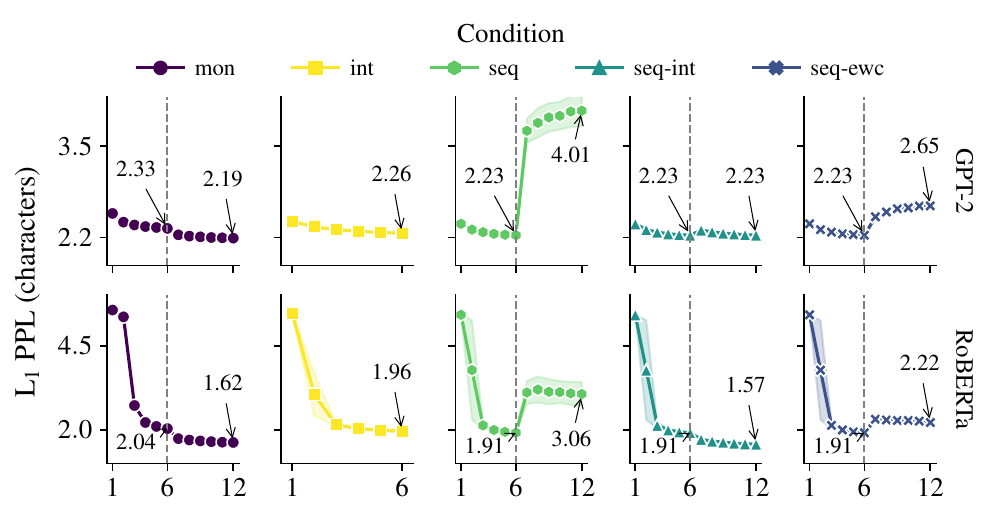}
        \label{fig:loss_l1_results}
    \end{subfigure}
    
     \vspace{-5pt}
     \begin{subfigure}[]{\columnwidth} 
          \centering
        \includegraphics[width=\columnwidth]{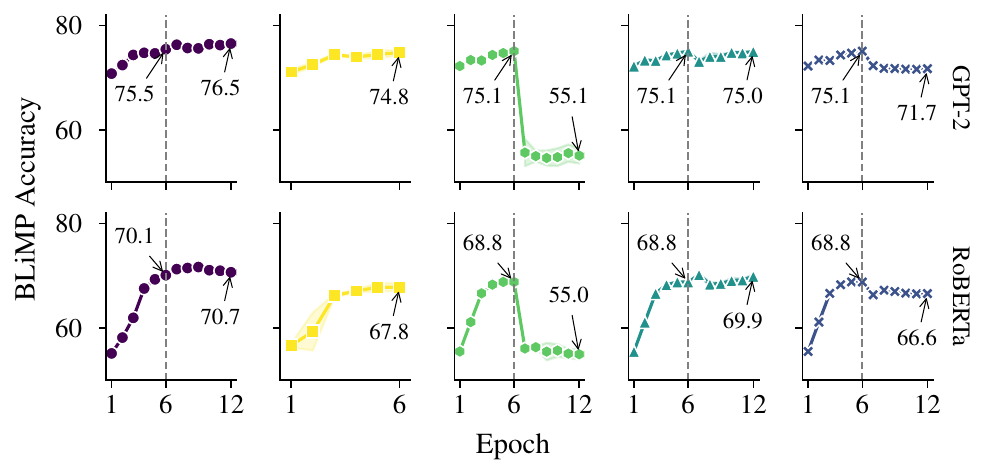}
        \label{fig:blimp_l1_results}
    \end{subfigure}
    
    \vspace{-10pt}
     \begin{subfigure}[]{\columnwidth}
        \centering
        \includegraphics[width=\columnwidth]{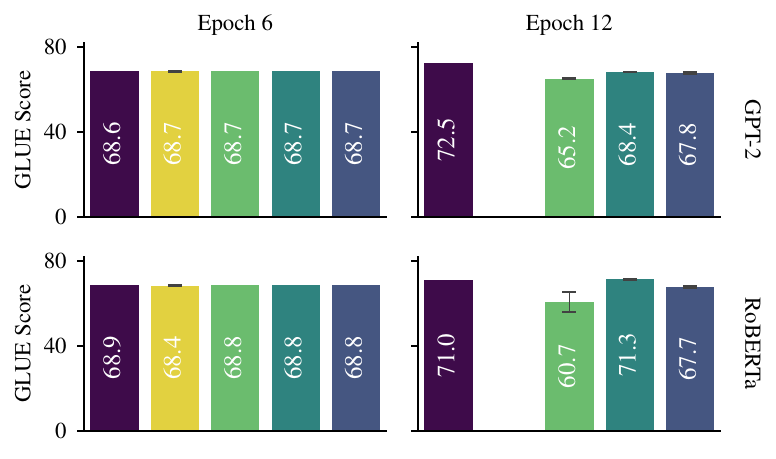}
        \label{fig:glue_l1_results}
    \end{subfigure}
    \vspace{-20pt}
    \caption{
    \lone ($\en$) results when the language order is reversed (6 + 6 epochs). Results are aggregated across \ltwo ($\de$ and $\fin$). 
    \textbf{Top:} \ppl per character on the \lone ($\en$) validation set during training.
    \textbf{Middle:} Accuracy on \blimp during training.
    \textbf{Bottom:} Performance on \glue at the end of training on \lone and \ltwo.
    }
    \label{fig:l1-results}
\vspace{-15pt}
\end{figure}

We illustrate the results for this experiment in \cref{fig:l2-results}. 
In general, the learning patterns through epochs are similar across conditions (except for \special{interleaved}), with variations observed mostly in the final performance. 
As expected, the \special{monolingual} performance is the best among all conditions.
We notice that the \special{interleaved} condition performs slightly worse than the \special{sequential} condition, although the difference is more noticeable in \roberta models than in \gpt models. Both achieve lower scores than the \special{monolingual} training (the baseline for native-level language proficiency).
Results from the \special{sequential-interleaved} condition differ based on the model architecture. On \gpt it is worse compared to \special{sequential} results (i.e., keeping \lone exposure hinders \ltwo  learning), while on \roberta it is better (i.e., keeping \lone exposure helps \ltwo learning).
Lastly, the \special{sequential-ewc} condition shows clear differences, with much worse \ltwo proficiency.
Unsurprisingly, the reduction in plasticity through regularization has a significant (negative) impact on \ltwo's learning outcome.

\subsection*{Experiment 2: Reversing the language order.\\
$\mathbf{\lone \in \{\en\}}$,
$\mathbf{\ltwo \in \{\de, \fin\}}$, 
$\mathbf{\size = 600M}$,
$\mathbf{\epochs = 6}$.} \label{sec:experiment-2}

\begin{figure*}[t]
    \begin{subfigure}[]{0.48\textwidth}
        \centering
        \includegraphics[width=\textwidth]{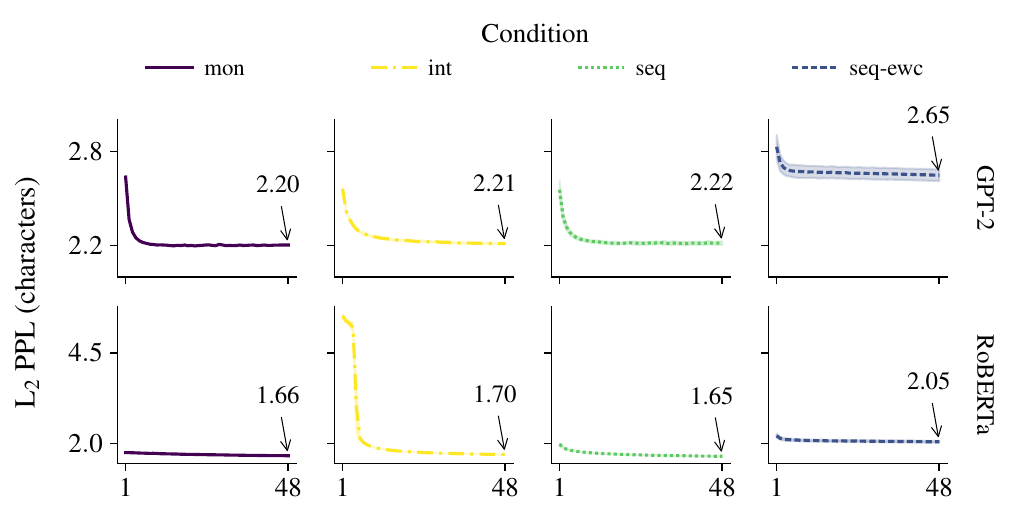}
        \label{fig:loss_l2_results_convergence}
    \end{subfigure}
    \hfill
    \begin{subfigure}[]{0.48\textwidth}
        \centering
        \includegraphics[width=\textwidth]{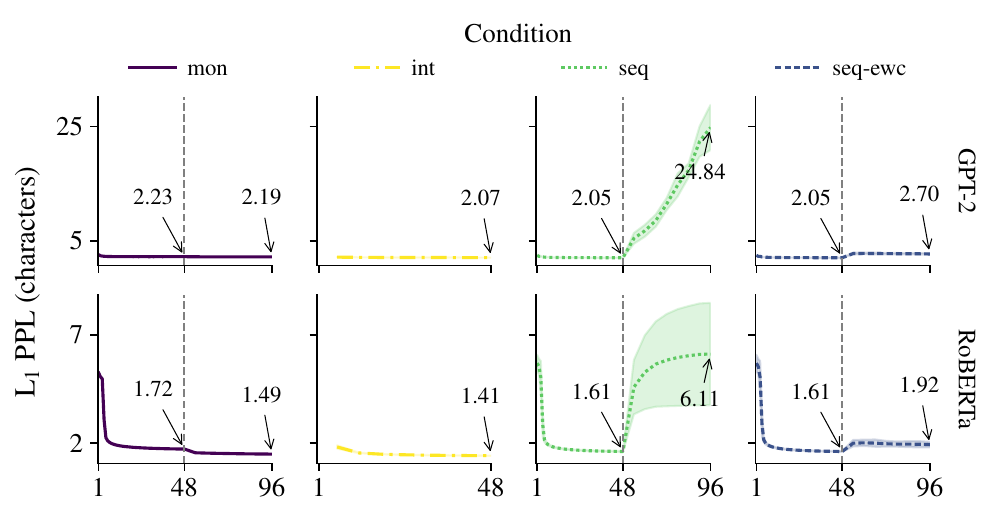}
        \label{fig:ppl_l1_results_convergence}
    \end{subfigure}  
    
    \vspace{-10pt}
     \begin{subfigure}[t]{0.48\textwidth}
        \centering
        \includegraphics[width=\textwidth]{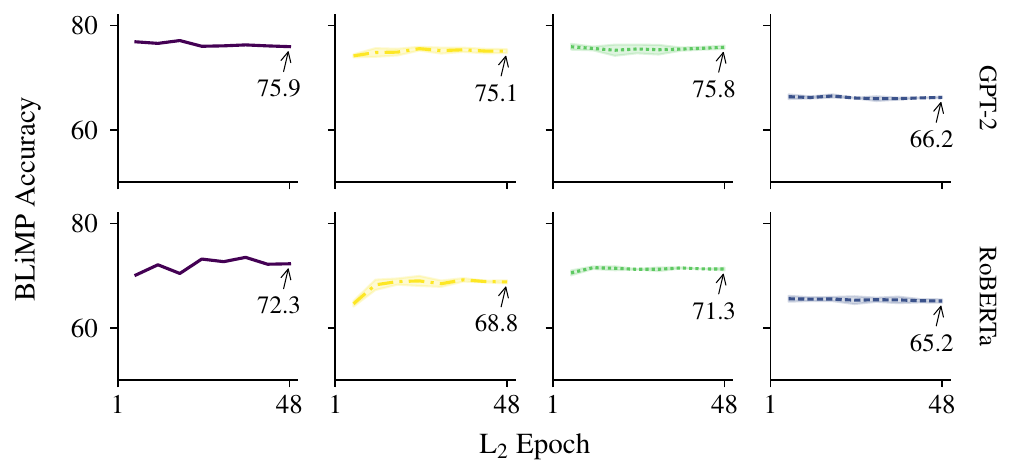}
        \label{fig:blimp_l2_results_convergence}
    \end{subfigure}
    \hfill
    \begin{subfigure}[b]{0.48\textwidth}
        \centering
        \includegraphics[width=\textwidth]{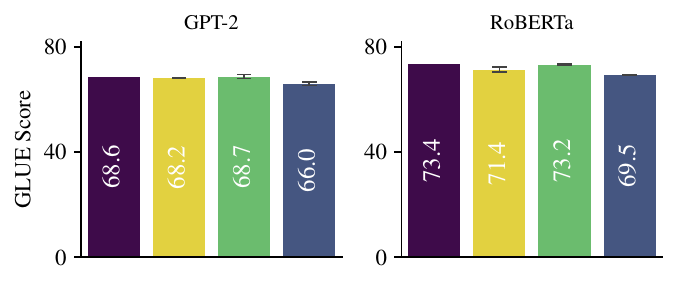}
        \label{fig:glue_l2_results_convergence}
    \end{subfigure}
    \vspace{-20pt}
    \caption{Summary of the \ltwo ($\en$) evaluation results for the convergence training (48 epochs). Results are aggregated across \lone ($\de$ and $\fin$). 
    \textbf{Top left:} \ppl per character on the \ltwo ($\en$) validation set during training on \ltwo.
    \textbf{Top right:} \ppl per character on the \lone ($\de$, $\fin$) validation set during training.
    \textbf{Bottom left:} Accuracy on \blimp during \ltwo training.
    \textbf{Bottom right:} Performance on \glue at the end of \ltwo training.
    }
    \label{fig:l2-results-convergence}
\vspace{-12.5pt}
\end{figure*}

The purpose of this experiment is to study the \cp for \lone attrition, which relates to \cref{hyp:cp_for_lone} and \cref{hyp:cp_with_ewc}.
The experimental setup differs from the previous one mainly in that the order of the languages is reversed: \texttt{English} is used as a first language and \texttt{German} and \texttt{Finnish} as second languages.
This swap allows us to use all three evaluation benchmarks to track \lone performance across the entire training process. 
Furthermore, each run is performed with a single configuration ($\mathbf{C_1}$).

The results for this experiment are displayed in \cref{fig:l1-results}. 
In general, we observe a smaller drop in \glue scores compared to \blimp scores after exposure to \lone ceases (i.e., comparing epoch 6 to 12).
This is most probably caused by the conceptual difference between evaluations in a zero-shot setting and evaluations that require fine-tuning. When fine-tuning on \glue, the \lm~is once again allowed to learn from \lone data. 
In the \special{monolingual} condition, \lone performance keeps improving until the end of training, even though it slows down in the second stage. 
In the \special{interleaved} condition, models reach the same level of \lone and \ltwo proficiency (when comparing these results to the ones from \cref{fig:l2-results}).\footnote{%
Note that the dataset size is 2\size in this condition, so the model is only trained for \epochs epochs.}
The more notable result comes from the \special{sequential} learning. 
In this condition, \lm s rapidly lose the knowledge acquired from \lone learning after \ltwo exposure is started. The final \lone perplexity values indicate that, in the end, the \lm s forget almost everything that they have learned before. 
It looks like the second language completely replaces the first one.
However, the loss of \lone is completely mitigated in the \special{sequential-interleaved} condition.
When \lone exposure is prolonged \emph{without} any reduction in plasticity, models are able to retain all the prior \lone knowledge.\footnote{\roberta continues learning \lone during the second stage. We believe this is due to undertraining.\looseness=-1}\looseness=-1

When a computational regularization method is introduced in the \special{sequential-ewc} condition, \lone knowledge is successfully preserved to a certain degree. 
\begin{figure}
    \centering
    \includegraphics[width=\columnwidth]{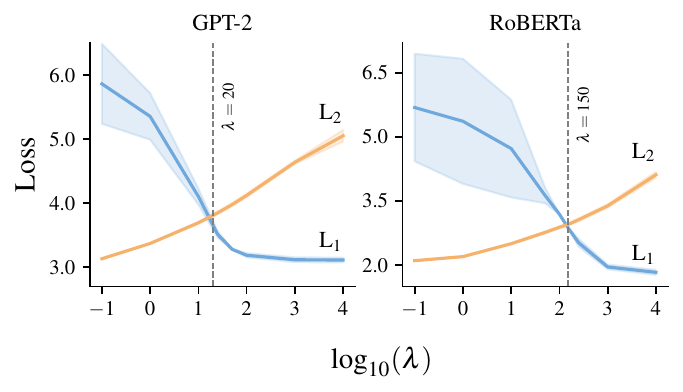}
    \vspace{-20pt}
   \caption{Trade-off between \lone and \ltwo performance (CE) at the end of training as a function of $\lambda$ (EWC strength). Results are aggregated across \lone ($\de$ and $\fin$).}
   \label{fig:ewc_lambda}
   \vspace{-15pt}
 \end{figure} 
However, as we have seen from \cref{fig:l2-results}, \ltwo learning is also harmed in this case. To explore this trade-off, we vary the $\lambda$ values and test the models' performance on both \lone and \ltwo at the end of training. The results are illustrated in \cref{fig:ewc_lambda}. We see that the regularization strength $\lambda$ highly influences both \lone and \ltwo learning outcomes. When high $\lambda$ values are used (strong \ewc regularization), \lone knowledge can be predictably maintained at the initial levels. However, \ltwo learning will be almost completely impaired. 
On the other side, when lower $\lambda$ values are used (weak \ewc regularization), \ltwo learning is affected less, but \lone will not be preserved. It is also noticeable that in this case, there is a higher variance on the final \lone outcomes. 
As mentioned in \cref{sec:ewc-implementation}, we have selected the $\lambda$ values for all our experiments as the point where \lone and \ltwo performance is roughly equivalent (intersection lines are marked on the plots). Thus, we do not favor either very strong or very weak regularization, and also to match the behavior exposed by the \special{interleaved} models.\looseness=-1

\begin{figure*}
    \begin{subfigure}[c]{\textwidth} 
        \centering
        \includegraphics[trim={0 1.7cm 0 0.25cm},clip,width=\textwidth]{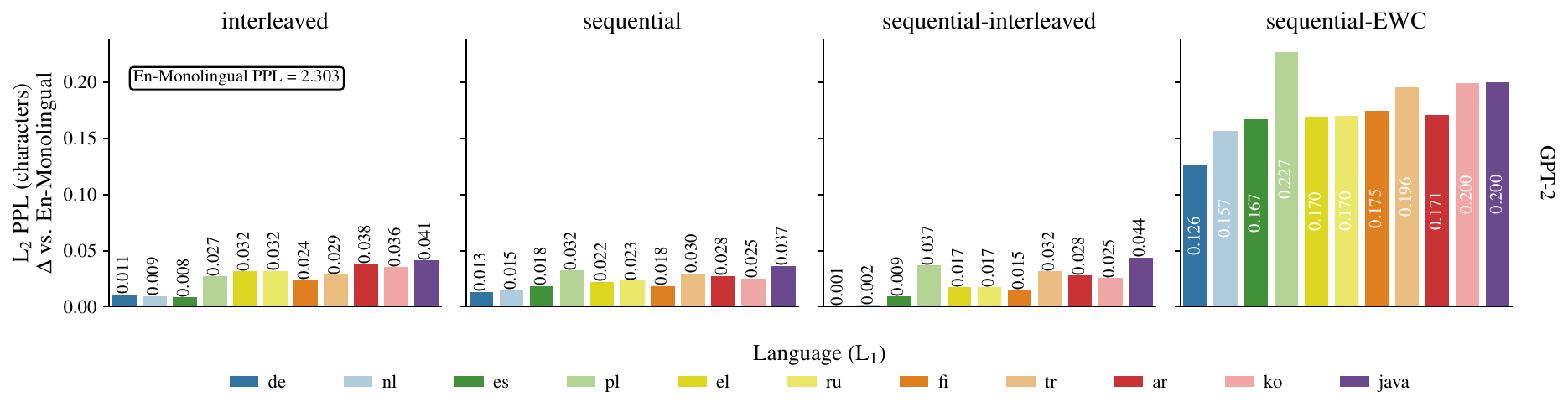}
        \label{fig:l2-ppl-multilingual}
    \end{subfigure}
        
    \begin{subfigure}[c]{\textwidth}
        \centering
        \includegraphics[trim={-0.8cm .4cm 0 0.7cm},clip,width=\textwidth]{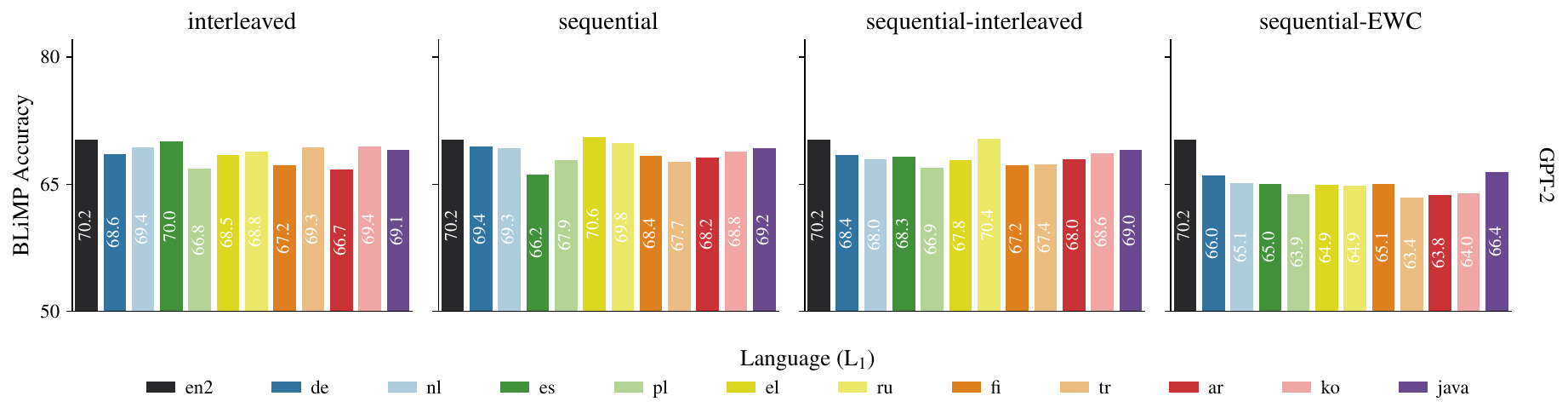}
        \label{fig:l2-blimp-multilingual}
    \end{subfigure}
    
    \caption{\ltwo ($\en$) results for each language pair at the end of training (6 epochs). 
    \textbf{Top:} \ppl per character on \ltwo ($\en$) at the end of training minus \ppl per character from the \special{monolingual} condition.
    \textbf{Bottom:} Accuracy on \blimp at the end of training.
    }
    \label{fig:l2-results-multilingual}
   \vspace{-15pt}
\end{figure*}

\subsection*{Experiment 3: Training to convergence.
$\mathbf{\lone \in \{\de, \fin\}}$, 
$\mathbf{\ltwo \in \{\en\}}$,
$\mathbf{\size = 600M}$,
$\mathbf{\epochs = 48}$.} \label{sec:experiment-3}

This experiment is motivated by the observation that \cp effects can become stronger with a later age of exposure.
We extend the training time for each language (thus postponing the start of \ltwo exposure) to 48 epochs, allowing the model weights to better converge.
As this is more computationally demanding, we do only one run per condition and we also omit the \special{sequential-interleaved} condition.

The results are provided in \cref{fig:l2-results-convergence}, which shows that the final results are more uniform across the first three conditions, especially between \special{interleaved} and \special{sequential}.
We still see a substantial loss in \lone performance in the \special{sequential} condition, indicating that longer \lone training does not lead to entrenchment, though these \ppl scores are for $\fin$ and $\de$, and thus are not directly comparable to the $\en$ from other experiments. 
The performance of the \special{interleaved} setting shows a slight improvement compared to regular training, indicating that a longer training duration was beneficial (and necessary for convergence). 
The learning pattern for the \special{sequential-ewc} condition does not change, i.e., the final \ltwo performance does not significantly improve with additional training. 
It also appears the \ewc regularization does not simply slow down \ltwo learning, but rather acts as a lower bound: \ltwo knowledge can never improve past a certain point for a given choice of $\lambda$.\looseness=-1

\subsection*{Experiment 4: Diversifying the language pool.
$\mathbf{\lone\!\in\!\{\ar, \de, \el, \es, \fin, \ko, \nl, \pl, \ru, \tr, \java\}}$, 
$\mathbf{\ltwo \in \{\en\}}$,
$\mathbf{\size = 100M}$,
$\mathbf{\epochs = 6}$.} \label{sec:experiment-4}

This experiment both addresses \cref{hyp:cp_language_pairs} and provides more diversity in the results; the latter is especially important considering the limited selection of languages for the previous experiments. 
The main focus here is to find whether the choice of languages or more concretely, the degree of similarity between \lone and \ltwo, has any impact on the \cp. 
For this, we consider a wider range of languages based on their relatedness to \texttt{English} (see \cref{sec:experimental-setup}). 
To accommodate for the increase in computational demand, we consider only the \gpt model architecture (with configuration $\mathbf{C_5}$), we reduce the size of the training data to 100 million tokens, we only train with English as \ltwo, and we only run \ppl and BLiMP evaluations.\looseness=-1

We present the findings in \cref{fig:l2-results-multilingual}.
Considering \ltwo PPL, we find the expected pattern, i.e., PPL generally increases (gets worse) when \lone is less closely related to it.
However, the effect size is quite small (hence we plot the difference with respect to \special{monolingual} condition): only 0.1-0.3 in most cases, while the total PPL ranges from about 2.3-2.5. 
However, the BLiMP results do not support the same conclusion.
Mostly, BLiMP scores vary seemingly at random, and indeed these differences could reflect random noise due to model initialization.
As expected, \ltwo performance on BLiMP \emph{is} greatest when \lone and \ltwo are just different corpora of English.
But curiously, \texttt{Java} pretraining aids BLiMP performance more than most natural languages.
The only condition where we can make the strongest case for an effect is \special{sequential-EWC}, which suggests that perhaps relatedness effects exist at the beginning of transfer but are wiped out by extensive \ltwo training.

\section{Discussion}

We first address our four research questions from \S \ref{sec:rqs}.
We then explore our results' implications for the theoretical claims introduced in \S \ref{sec:role_of_lms_in_cp}.\looseness=-1

\subsection{Research Questions}

\paragraph{\cref{hyp:cp_for_ltwo}.} 
This RQ concerned \ltwo \cp effects.
We find evidence against a \cp for \ltwo learning in language models. 
In experiments 1 and 3 (\cref{fig:l2-results,fig:l2-results-convergence}), we consistently find that final performance in \ltwo is \emph{worse} (or at least not different) in the \special{interleaved} than in the \special{sequential} condition.
This is the opposite of the pattern found in humans, where bilingual learners have greater proficiency in \ltwo with earlier exposure to it \citep{johnson1989critical}.
Comparing the \special{interleaved} and \special{monolingual} conditions also informs this question.
For humans, early \ltwo learners resemble monolinguals in their \ltwo performance. 
However, \special{monolingual} models outperform \special{interleaved} ones on all three metrics.
Looking at the entire learning trajectory, \special{interleaved} models differ markedly from both \special{monolingual} and \special{sequential} models, showing more gradual and delayed improvements.\looseness=-1

\paragraph{\cref{hyp:cp_for_lone}.} This RQ concerned the \cp for \lone attrition.
Humans show few signs of \lone attrition after the \cp, even if \lone exposure decreases or ceases.
Our results in experiment 2 provide strong evidence against a similar phenomenon in typical \lm s.
Instead, we find that \lone performance worsens rapidly and to a large degree in the \special{sequential} condition after \lone exposure ceases.
This is expected given the susceptibility of neural networks to catastrophic forgetting. 
The loss of \lone proficiency is prevented by continuing \lone exposure in the \special{sequential-interleaved} condition, but such continued exposure is not necessary in humans.\looseness=-1
\footnote{This ignores self-talk as a potential source of \lone exposure; self-talk may continue even if external \lone exposure ceases.}\looseness=-1

\paragraph{\cref{hyp:cp_with_ewc}.} This RQ concerned the trade-off between \ltwo learning and preventing \lone attrition when explicitly reducing plasticity partway during learning.
We find strong evidence for such a trade-off when comparing the \special{sequential-EWC} condition to the \special{sequential} condition.
The value of $\lambda$ we selected preserved \lone performance substantially compared to the \special{sequential} models, at the cost of harming final performance in \ltwo.
The \ltwo learning curves also converge relatively quickly when plasticity is reduced.
Our exploration of different values of $\lambda$ (\cref{fig:ewc_lambda}) shows that preserving \lone to monolingual levels harms \ltwo acquisition by roughly 1.5 nats of \ltwo performance, but we do not directly compare grammatical performance of our models to that of human late \ltwo learners.
Thus, our attempt at reverse engineering shows a broadly human-like learning pattern when using EWC, but we cannot say quantitatively and at a high level of granularity whether the result is human-like.\looseness=-1

\paragraph{\cref{hyp:cp_language_pairs}.} 
This RQ concerned the impact of language similarity on \cp effects.
Our results showed that the language family of \lone and its script has an impact on \ltwo learning in the expected way for only a subset of evaluations.
Based on earlier findings from \citet{papadimitriou-jurafsky-2020-learning} and \citet{oba2023second}, we had expected \ltwo performance would be greater when \lone is more closely related.
However, only our results for \ppl support this prior conclusion.
Our results for BLiMP do not unless EWC is applied, suggesting that models are ordinarily plastic enough to learn the grammar of \ltwo regardless of relatedness, unless plasticity is specifically reduced.
We note that \citet{papadimitriou-jurafsky-2020-learning} reduce plasticity by freezing the model weights before transferring to \ltwo, but we do not venture an explanation why \citet{oba2023second} seemingly find models to be less plastic than we do.\looseness=-1 

\subsection{Theoretical Implications}

Our results show that \cp effects are not naturally arising in \lm s in a typical training regime.
At the same time, we are able to suggest a methodology to reverse engineer human-like learning patterns by artificially reducing plasticity later in training.
As discussed in Section \ref{sec:role_of_lms_in_cp}, results like these are relevant to certain specific claims in the critical period literature.
Specifically, they refute the \strongExperience, which states that all successful learning algorithms will show \cp effects, and they are consistent with (but do not provide strong evidence for) the \strongInnate, which states that innate maturational stages are necessary to produce \cp effects.\looseness=-1

The \strongExperience incorrectly predicts that our \lm s will naturally show \cp effects.
This may come as a surprise, given this claim was based on previous studies on connectionist models finding evidence for the phenomenon of entrenchment \citep{munro1986statedependent,ellis2000age}.
One explanation for this discrepancy in results is the difference in models' capacity.
These earlier works trained extremely small models by today's standards, whereas our \lm s may be over-parameterized and therefore have sufficient capacity to train to convergence without entrenchment.

On the other hand, the \strongInnate correctly predicts that our \lm s will not show \cp effects, unless we introduce an innate loss in plasticity.
Our introduction of EWC part-way through training is akin to an innate loss in plasticity.
However, our experiments are not strong evidence either that the \strongInnate is correct or that EWC is a plausible algorithmic mechanism underlying the loss in plasticity in humans.
The possibilities remain that other statistical learners will show \cp effects as a natural consequence of experience, and that humans could be one such learner. 

Finally, many scholars of human development advocate for a complex explanation of \cp effects involving both innate and experiential mechanisms \citep[e.g.,~][]{newport1990maturationala,thiessen2016statistical,singleton2005critical}.
We consider this nuanced view to be likely correct, but our results suggest that the role of statistical learning should not be assumed or overstated without evidence from humans or more cognitively plausible models.

\subsection{Limitations and Future Work}

From a Bayesian epistemological point of view, results from GPT-2 and RoBERTa should affect our priors about general learners and humans in \emph{some} ways, but Transformer LMs are inherently limited as models of human learners.
Assuming a sort of Copernican Principle for cognitive modeling, humans and LMs should both be unextraordinary relative to the theoretical class of language learners. 
So without other evidence or \emph{a priori} reasoning, we should assume they share properties: i.e., the property that we identified in LMs that the ordinary mechanism of learning fails to lead to \cp effects.
However, there are many differences that could lead us \emph{a priori} to hypothesize differences in how humans and LMs learn.
More generalizable evidence could be obtained by considering more cognitively-inspired models and learning algorithms, and learning environments drawing on more developmentally plausible linguistic data and multimodal input \citep{warstadt2022what}.
For those who want to defend the position that \cp effects fall out in humans---but not LMs---from ordinary learning, these or other differences must be identified as the cause.
Our experiments seek a compromise given today's resources by choosing architectures and procedures that maximize human-like learning \emph{outcomes}, while still using a transcribed-speech training corpus and human-scale data.
Future work should revisit this compromise as better resources for cognitive modeling are developed.\looseness=-1

One caveat to the rejection of \strongExperience is that the claim only applies to learners that adequately acquire \lone.
While this condition is not rigorously defined, one might argue our models do not qualify.
For example, our GPT-2 models in the \special{sequential} and \special{interleaved} conditions achieve $75\pm 1\%$ accuracy on BLiMP.
This is substantially better than chance (50\%) and is comparable to the strongest baseline model from the BabyLM Challenge \citep{warstadt2023findings}, but falls well below human agreement with BLiMP (89\%).
No \lm s currently achieve fully human-like performance on BLiMP, and higher performance generally requires compromising on developmental plausibility, though more data-efficient architectures exist \citep{samuel2023trainedb,warstadt2023findings}.
We do not expect our results to be qualitatively different if our experiments are run with more effective Transformer-based \lm s, but this is something future work should confirm.

We must also acknowledge the theoretical implications of learning rate decay and the optimizer.
The learning rate impacts the magnitude of changes to model weights during training, and so it is directly related to the plasticity of the model. 
Extensive work in machine learning has found that the learning rate should decrease in the later stages of training \citep[see, e.g.,][]{gotmare2018a}.
Thus, one could take this as evidence that a predetermined loss in plasticity is necessary for successful learning, though the implementation may not necessarily result in \cp effects.
As our goal with our experiments was to reproduce typical \lm{} training pipelines, we reduce the learning rate in all conditions, but we also restart the learning rate in the sequential conditions at the beginning of \ltwo training, as is standard in fine-tuning \citep{howard2018universal}.
However, this may be interpreted as artificially \emph{increasing} the plasticity of our models, which could contribute to the lack of \cp effects.
Learning rate schedules, their interaction with successful learning, and their impact on critical periods should be the focus of future work.

We identify several additional avenues for future work:
First, the regularization method we use, EWC \citep{kirkpatrick2017overcoming}, can be viewed only as a computational-level model of an innate biological \cp; future work could consider other regularizers such as Memory Aware Synapsis \citep{aljundi2018memory} that arguably model what happens in humans at the algorithmic level.
Second, there is neurolinguistic evidence for some degree of modularity in how human bilinguals process different languages \citep{hernandez2005emergence}, but our models use completely shared parameters for languages.
Future work can explore \lm{} architectures that encourage or directly build in modularity, such as XLM \citep{lample2019cross} or X-MOD \citep{pfeiffer-etal-2022-lifting}.
Third, our models learn in a text-only environment, but for humans \lone and \ltwo are both grounded in the same non-linguistic stimuli.
Training multimodal models can lead to more realistic simulations, as well as enable testing of \cp effects for \lone learning, which requires non-linguistic experience to precede \lone acquisition.

\section{Conclusion}

There are many obvious ways in which \lm s differ from humans when learning language.
Our work reveals another important point of divergence, namely that \lm s remain far more plastic later into the learning process than human learners.
Even though humans and models differ substantially, comparing the learning trajectories of human learners to those of computational ones tells us something about humans:
Those features that we do share are more likely to be natural properties of language learning, while those that we do not are more likely to require idiosyncratic innate mechanisms.
Our results provide strong evidence against the hypothesis that \cp effects are necessarily induced solely by experience, and they are consistent with, but only provide weak evidence in favor of, the view that innate mechanisms are necessary to explain \cp phenomena.
It will be important to replicate these results in other artificial learners, including ones which resemble humans more closely in other respects.
Our study thus constitutes early progress towards integrating modern \lm s into the study of human language acquisition.

\section*{Acknowledgments}
We thank the editors and reviewers whose feedback has led to substantial improvements in this work.
TP and AW are supported by ETH Postdoctoral Fellowships.

\bibliography{anthology,custom}
\bibliographystyle{acl_natbib}

\appendix

\clearpage

\section{Data Preprocessing} \label{app:data_preprocess}

\paragraph{Cleaning.}
We remove extra spaces, non-breaking spaces and dataset-specific characters such as dialogue lines and music symbols (in \subtitles) and paragraph delimiters (in \gutenberg). For the \texttt{Java} code extracted from \thestack, we remove all docstrings and comments.

\paragraph{Unifying.}

We downsample the original data sources to align the data quantity and distribution of domains for all languages. A fixed sampling ratio of $2:1:1$ is established for $\subtitles:\gutenberg:\wikipedia$ in order to ensure an even distribution of transcribed and written text within the training data. 
As mentioned previously, there is an exception to this rule for Experiment 4, where only data from \subtitles and \wikipedia is sampled in equal parts. 
To mix the data from different domains while limiting the number of context breaks, large blocks of $10000$ lines are uniformly sampled from each dataset, and then randomly shuffled. 
The resulting unified dataset is then split as follows: $83\%$ train, $8.5\%$ validation, and $8.5\%$ test.

\paragraph{Interleaving.} \label{subsec:interleaving}

Interleaved datasets are created for each language pair ($\en$ plus a second language). The same blocks of texts sampled in the previous unifying step are simply interleaved while maintaining their ordering (e.g., \lone block 1, \ltwo block 1, \lone block 2, \ltwo block 2, etc.). In this way, the data from the two languages is presented in the same order to the model during the \special{interleaved} training as it is during the \special{sequential} training.

\paragraph{Size alignment.}
We uniformize the training dataset sizes across languages. 
We quantify dataset sizes in terms of the number of tokens obtained using a \bpe tokenizer; see section \cref{sec:tokenization}.
Beyond the fact that tokens are the true unit of input to the \lm s, \bpe is a compression algorithm, so while the amount of information per word might be highly language-specific, the amount of information per token is more comparable across languages.

\paragraph{Tokenization.}\label{sec:tokenization}

We train bilingual byte-Level BPE tokenizers.\footnote{With the exception of the \special{monolingual} condition, which uses a monolingual byte-Level BPE tokenizer instead.}
Given the size of our training datasets, we set a vocabulary size of 32,000 and a minimum frequency of 2. 
When tokenizing the training data, all the text lines are concatenated and the resulting tokenized dataset is split into fixed-size blocks of 512 tokens which are used as input for both \gpt and \roberta.

\section{Data}\label{app:data}

\begin{table}[ht]
\centering
\resizebox{.95\linewidth}{!}{%
\begin{tabular}{lrrrr}
\toprule
\textbf{Dataset} & \textbf{Lang.} & \textbf{Size (GB)} & \textbf{Lines (M)} & \textbf{Words (M)} \\
\midrule
\multirow{11}{*}{\subtitles} & $\ar$ & 1.6 & 39 & 177 \\
 & $\de$ & 1.2 & 34 & 202 \\
 & $\el$ & 6.5 & 126 & 650 \\
 & $\en$ & 11.0 & 316 & 2112 \\
 & $\es$ & 6.4 & 213 & 1144 \\
 & $\fin$ & 1.4 & 45 & 191 \\
 & $\ko$ & 0.1 & 3 & 8 \\
 & $\nl$ & 3.1 & 105 & 600 \\
 & $\pl$ & 6.8 & 236 & 1055 \\
 & $\ru$ & 2.2 & 45 & 214 \\
 & $\tr$ & 5.1 & 173 & 698 \\
\midrule
\multirow{11}{*}{\wikipedia} & $\ar$ & 2.1 & 8 & 209 \\
 & $\de$ & 6.4 & 25 & 931 \\
 & $\el$ & 1.1 & 2 & 92 \\
 & $\en$ & 15.0 & 60 & 2543 \\
 & $\es$ & 4.5 & 21 & 767 \\
 & $\fin$ & 0.8 & 3 & 93 \\
 & $\ko$ & 0.8 & 5 & 85 \\
 & $\nl$ & 1.9 & 12 & 303 \\
 & $\pl$ & 2.0 & 11 & 275 \\
 & $\ru$ & 7.4 & 24 & 604 \\
 & $\tr$ & 0.7 & 4 & 86 \\
\midrule
\multirow{3}{*}{\gutenberg} & $\en$ & 19.0 & 59 & 3417 \\
 & $\de$ & 0.6 & 2 & 103 \\
 & $\fin$ & 0.7 & 3 & 92 \\
\midrule
\thestack & $\java$ & 3.8 & 110 & 337 \\
\bottomrule
\end{tabular}%
}
\caption{Statistics of the collected data}
\label{tab:data-statistics}
\end{table}

\begin{table}[h!]
    \centering
    \resizebox{0.95\linewidth}{!}{%
    \begin{tabular}{lccc}
    \toprule
    \textbf{Train   Dataset} & \textbf{Size (GB)} & \textbf{Words (M)} & \textbf{Tokens (M)}\\
    \midrule
    $\en$ & 2.2 & 408 & 601\\
    $\en2$ & 2.2 & 402 & 592\\
    $\de$ & 2.4 & 378 & 600\\
    $\fin$ & 2.4 & 311 & 596\\
    \bottomrule
    \end{tabular}%
    }
    \caption{Main experiments' dataset sizes}
    \label{tab:unified-clean-counts}
\end{table}

\begin{table}[h!]
    \centering
    \resizebox{0.95\linewidth}{!}{%
    \begin{tabular}{lccc}
    \toprule
    \textbf{Train Dataset} & \textbf{Size (GB)} & 
    \textbf{Words (M)} & \textbf{Tokens (M)} \\
    \midrule
    $\ar$ & 0.6 & 62 & 104 \\
    $\de$ & 0.4 & 63 & 100\\
    $\el$ & 0.7 & 61 & 102\\
    $\en$ & 0.4 & 68 & 99\\
    $\en2$ & 0.4 & 66 & 96\\
    $\es$ & 0.4 & 66 & 100\\
    $\fin$ & 0.4 & 51 & 100\\
    $\ko$ & 0.5 & 47 & 102\\
    $\nl$ & 0.4 & 65 & 98\\
    $\pl$ & 0.4 & 57 & 104\\
    $\ru$ & 0.6 & 54 & 99\\
    $\tr$ & 0.4 & 55 & 101\\
    $\java$ & 0.3 & 30 & 99\\
    \bottomrule
    \end{tabular}%
    }
    \caption{Experiment 4 dataset sizes}
    \label{tab:unified-extended-counts}
\end{table}

\clearpage
\section{Hyperparameters}\label{app:hyperparams}

\begin{table}[h!]
\centering
\resizebox{\linewidth}{!}{%
\renewcommand{\arraystretch}{1.5}%
\begin{tabular}{lccccc}
\toprule
     \textbf{Hyperarameter}&   $\mathbf{C_1}$ & $\mathbf{C_2}$ & $\mathbf{C_3}$ & $\mathbf{C_4}$ & $\mathbf{C_5}$\\
     \midrule
     Learning rate ($\times10^{-3}$)&  $1.00$& $1.00$& $8.00$ & $1.00$ & $1.00$\\
    Warmup ratio & $7\%$& $9\%$& $10\%$ & $7\%$ & $7\%$\\
    Gradient accum. steps&  $16$& $32$& $32$ & $16$ & $4$\\
\bottomrule
\end{tabular}%
}
\caption{Configurations for \gpt models.}
\label{tab:gpt2-config}
\end{table}

\begin{table}[h!]
\centering
\resizebox{\linewidth}{!}{%
\renewcommand{\arraystretch}{1.5}%
\begin{tabular}{lccccc}
\toprule
     \textbf{Hyperarameter}&  $\mathbf{C_1}$ & $\mathbf{C_2}$ & $\mathbf{C_3}$ & $\mathbf{C_4}$ & $\mathbf{C_5}$\\
     \midrule
     Learning rate ($\times10^{-4}$)&  $4.75$ & $3.88$& $3.90$ & $3.00$ & $4.00$\\
    Warmup ratio& $5\%$& $10\%$& $9\%$ & $1\%$ & $1\%$\\
    Gradient accum. steps&  $32$& $32$& $16$ & $32$ & $1$\\
    MLM probability& $0.3$& $0.15$& $0.3$ & $0.3$ & $0.3$\\
\bottomrule
\end{tabular}%
}
\caption{Configurations for \roberta models.}
\label{tab:roberta-config}
\end{table}

\begin{table}[ht]
        \centering
        \resizebox{\linewidth}{!}{%
        \begin{tabular}{lrr}
        \toprule
        & \multicolumn{2}{c}{\textbf{Value}} \\\cmidrule(lr){2-3}
        \textbf{Hyperparameter} & \textbf{\gpt} & \textbf{\roberta} \\
        \midrule
        n\_head & 12 
        &
        12 \\
        n\_layer & 12 
        &
        12 \\
        n\_positions & 1,024 
        &
        512 \\
        n\_embd & 768 
        &
        768 \\
        activation\_function & ``gelu\_new'' 
        &
        ``gelu'' \\
        optimizer & ``adamw\_hf'' 
        &
         ``adamw\_hf'' \\
        lr\_scheduler & ``linear'' 
        &
        ``linear'' \\
        device\_train\_batch\_size & 4 
        &
        8 \\
        adafactor & False
        &
         False \\
        adam\_beta1 & 0.9 
        &
        0.9 \\
        adam\_beta2 & 0.999 
        &
        0.999 \\
        adam\_epsilon & 0.00000001 
        &
        0.00000001 \\
        max\_grad\_norm & 1 
        &
        1 \\
        layer\_norm\_epsilon & 0.00001 
        &
        0.000000000001 \\
        weight\_decay & 0 
        &
        0 \\
        dropout & 0.1 
        &
         0.1 \\
        fp16 & True 
        &
         True \\
        \bottomrule
        \end{tabular}%
        }
       \caption{Fixed hyperparameters for \gpt and \roberta models.}
       \label{tab:gpt2-params}
\end{table}

\newpage
\section{Evaluation Tasks}\label{app:eval}

\begin{table*}[ht]
\resizebox{\textwidth}{!}{%
\renewcommand{\arraystretch}{1.7}%
\begin{tabular}{clll}
\toprule
\textbf{Field} & \textbf{Phenomenon} & \textbf{Acceptable example} & \textbf{Unacceptable example} \\
 \midrule
 \multirow{4}{*}{\textbf{Morphology}} 
 & \textsl{Anaphor Agreement} & \textit{Many girls insulted \underline{themselves}.} & \textit{Many girls insulted \underline{herself}.} \\
 & \textsl{Determiner-Noun Agreement} & \textit{Rachelle had bought that \underline{chair}.} & \textit{Rachelle had bought that \underline{chairs}.} \\
 & \textsl{Irregular Forms} & \textit{Aaron \underline{broke} the unicycle.} & \textit{Aaron \underline{broken} the unicycle.} \\
 & \textsl{Subject-Verb Agreement} & \textit{These casseroles \underline{disgust} Kayla.} & \textit{These casseroles \underline{disgusts} Kayla.} \\
 \midrule
 \multirow{5}{*}{\textbf{Syntax}} & \textsl{Argument Structure} & \textit{Rose wasn’t \underline{disturbing} Mark.} & \textit{Rose wasn’t \underline{boasting} Mark.} \\
 & \textsl{Ellipsis} & \begin{minipage}[t]{0.6\columnwidth}%
\textit{Jill hides one \underline{orange} chair and Tammy hides more.} 
\end{minipage} & \begin{minipage}[t]{0.6\columnwidth}%
\textit{Jill hides one chair and Tammy hides more \underline{orange}.}
\end{minipage}\\
 & \textsl{Filler-Gap} & \textit{Brett knew \underline{what} many waiters find.} & \textit{Brett knew \underline{that} many waiters find.} \\
 & \textsl{Island Effects} & \textit{Which \underline{bikes} is John fixing?} & \textit{Which is John fixing \underline{bikes}?} \\
 & \textsl{Subject-Auxiliary Inversion} & \textit{\underline{Was} the steak he \underline{is} cooking fresh?}  & \textit{\underline{Is} the steak he cooking \underline{was} fresh?} \\
 \midrule
 \multirow{3}{*}{\textbf{Semantics}} 
 & \textsl{NPI Licensing} & \textit{The truck has \underline{clearly} tipped over.} & \textit{The truck has \underline{ever} tipped over.} \\
 & \textsl{Quantifiers} & \textit{No boy knew \underline{fewer than} six guys.} & \textit{No boy knew \underline{at most} six guys.} \\
 & \textsl{Hypernym} & \textit{He has a chihuahua, so he has a \underline{dog}.} & \textit{He has a chihuahua, so he has a \underline{cat}.} \\
 \midrule
 \multirow{2}{*}{\textbf{Syntax \&   Semantics}} & \textsl{Binding} & \textit{Carlos said that Lori helped \underline{him}.} & \textit{Carlos said that Lori helped \underline{himself}.} \\
 & \textsl{Control/Raising} & \textit{There was \underline{bound} to be a fish escaping.} & \textit{There was \underline{unable} to be a fish escaping.} \\
 \midrule
 \multirow{3}{*}{\textbf{Discourse}} 
 & \textsl{Q-A Congruence (easy)} & \textit{A: Who is sleeping? B: \underline{David}.} & \textit{A: Who is sleeping? B: \underline{Eggs}.}  \\
 & \textsl{Q-A Congruence (tricky)} & \textit{A: Who studies? B: \underline{David}.} &  \textit{A: Who studies? B: \underline{Science}.} \\
 & \textsl{Turn-taking} & \textit{A: Did you arrive? B: No, \underline{we} didn't.}  & \textit{A: Did you arrive? B: No, \underline{you} didn't.}\\
 \bottomrule
\end{tabular}%
}
\caption{Examples of \blimp minimal pairs \citep{warstadt-etal-2020-blimp-benchmark}.}
\label{tab:blimp-examples}
\end{table*}

The BLiMP tasks that our models are evaluated on are examplified in \cref{tab:blimp-examples}.
The subset of \glue\footnote{\url{https://gluebenchmark.com}} tasks that our models are evaluated on are the following.

    \paragraph{CoLA.} The Corpus of Linguistic Acceptability consists of around 10K English sentences gathered from published linguistics literature that have been annotated by experts for binary acceptability (grammaticality) judgements \citep{warstadt2018neural}. The evaluation metric is Matthews Correlation Coefficient.\looseness=-1 
    \paragraph{SST-2.} The Binary Stanford Sentiment Treebank includes 215K unique phrases extracted from the parse trees of movie reviews sentences and that have been fully labeled as having either a positive or negative sentiment by three human judges \citep{socher2013recursive}. The evaluation metric is Accuracy.\looseness=-1
    \paragraph{MRPC.} The Microsoft Research Paraphrase Corpus contains 5,800 pairs of sentences sourced from a large corpus of news data, each labeled with a binary judgment indicating whether the pair represents a paraphrase or not \citep{dolan2005automatically}. The evaluation metric is F1 score.
    \paragraph{QQP.} The Quora Question Pairs is a corpus of over 400K question pairs from the Quora website which are annotated with a binary value denoting whether the questions are paraphrases of each other. The evaluation metric is F1 score.
    \paragraph{MNLI.} The Multi-Genre Natural Language Inference is a crowd-sourced collection of 433K sentence pairs annotated with textual entailment information, covering a wide range of genres of both spoken and written text  \citep{williams2018broad}. The evaluation metric is Accuracy.
    \paragraph{MNLI-MM.} The Multi-Genre Natural Language Inference Mismatched dataset is the mismatched version of the MNLI (matched) dataset, where the dev and test sets use out-of-domain data that does not closely resemble anything seen at training time \citep{williams2018broad}. The evaluation metric is Accuracy.
    \paragraph{QNLI.} The Question-answering Natural Language Inference dataset is automatically derived from the Stanford Question Answering Dataset v1.1 (SQuAD) \citep{rajpurkar2016squad}. It consists of question--paragraph pairs with one sentence in each paragraph, sourced from Wikipedia, containing the answer to the corresponding question, written by an annotator \citep{wang2019glue}. The evaluation metric is Accuracy.
    \paragraph{RTE.} The Recognizing Textual Entailment dataset is compiled from a series of textual entailment challenges: RTE1 \citep{dagan2006pascal}, RTE2 \citep{bar2006second}, RTE3 \citep{giampiccolo2007third}, and RTE5 \citep{bentivogli2009fifth}. The task requires to recognize whether the meaning of a text fragment can be inferred from the other text. The evaluation metric is Accuracy.\\

\noindent The subset of \sglue\footnote{\url{https://super.gluebenchmark.com}} tasks that the models are evaluated on are the following.
    \paragraph{BoolQ.} The Boolean Questions dataset is a reading comprehension dataset that consists of almost 16K naturally occurring yes--no questions generated in unprompted and unconstrained settings \citep{clark2019boolq}. The evaluation metric is Accuracy.
    \paragraph{MultiRC.} The Multi-Sentence Reading Comprehension dataset contains around 10K questions that can be answered by combining information from a multi-sentence paragraph \citep{khashabi2018looking}. The evaluation metric is F1 score.
    \paragraph{WSC.} The Winograd Schema Challenge dataset is a corpus of sentence pairs that differ in only one or two words and contain an ambiguity that can be resolved using world knowledge and reasoning \citep{levesque2011winograd}. The evaluation metric is Accuracy.

\newpage
\onecolumn 
\section{Derivation of \ewc Regularization for Language Modeling} \label{app:ewc-derivation}
\newcommand{\eos}{\textsc{eos}}
Let $\Sigma$ be an alphabet; furthermore, define $\overline{\Sigma} \defeq \Sigma \cup \{\eos\}$.
We assume we have access to a collection of strings $\dataone = \{\strone^{(n)}\}_{n=1}^N \subset \Sigma^*$ in \lone and another $\datatwo = \{\strtwo^{(m)}\}_{m=1}^M \subset \Sigma^*$ in \ltwo.
Additionally, let $\parameters \subset \R^d$ be a compact set of possible parameters.
We take a Bayesian approach and construct a posterior density over possible parameter vectors $\vtheta$.
Let $\prior(\vtheta)$ be a prior density over $\parameters$, and consider the following likelihood
\begin{subequations}\label{eq:likelihood}
\begin{align}
p(\dataone, &\datatwo \mid \vtheta) = \underbrace{\prod_{n=1}^N p(\strone^{(n)} \mid \vtheta)}_{\defeq p(\dataone \mid \vtheta)} \underbrace{\prod_{m=1}^M p(\strtwo^{(m)} \mid \vtheta)}_{\defeq p(\datatwo \mid \vtheta)} \\
&= \prod_{n=1}^N \prod_{t=1}^{T_n} p(\eos \mid \strone^{(n)}, \vtheta) p(x_{t}^{(n)} \mid \strone_{<t}^{(n)}, \vtheta) \prod_{m=1}^M \prod_{t=1}^{T_m} p(\eos\mid \strtwo^{(m)}, \vtheta) p(y_t^{(m)} \mid \strtwo_{<t}^{(m)}, \vtheta)
,
\end{align}
\end{subequations}
where, as the notation states, our model assumes conditional independence between data instances given the model's parameters.
Note that our language model $p(\cdot \mid \vtheta)$ is \emph{unusual} in that it generates sentences from both $\lone$ and $\ltwo$.
By Bayes' rule, we have the following posterior
\begin{subequations}
\begin{align}
    p(\vtheta \mid \dataone, \datatwo) &\propto p(\dataone, \datatwo \mid \vtheta) \prior(\vtheta) \label{eq:post1}\\
    &= p(\datatwo \mid \vtheta) p(\dataone \mid \vtheta) \prior(\vtheta)  \label{eq:post2} \\
    &\propto p(\datatwo \mid \vtheta) p(\vtheta \mid \dataone),
\end{align}
\end{subequations}
where the transition from \cref{eq:post1} to \cref{eq:post2} follows from the conditional independence assumption of \cref{eq:likelihood}.
There is no known general-purpose algorithm to compute $p(\vtheta \mid \dataone, \datatwo)$.
Thus, we construct a Gaussian approximation of $p(\vtheta \mid \dataone)$ by computing a second-order Taylor approximation to the log-likelihood 
around the likelihood mode, i.e.,
\begin{equation}
    \thetaone = \argmax_{\vtheta \in \parameters} \log p(\dataone \mid \vtheta).
\end{equation}
Applying the Taylor approximation yields
\begin{equation} \label{eq:taylor-approx}
    \log\, p(\dataone \mid \vtheta) 
    \approx \log p(\dataone \mid \thetaone) + \frac{1}{2}  (\vtheta - \thetaone)^\intercal \nabla^2_{\vtheta} \log p(\dataone \mid \thetaone)  (\vtheta - \thetaone).
\end{equation}
We have omitted the first term in the Taylor expansion since it is zero precisely because we have expanded $\log p(\dataone \mid \vtheta)$ around a local optimum.
If the prior is set to be a zero-centered, spherical Gaussian with variance $\sigma^2$, i.e.,
\begin{equation}\label{eq:prior}
    \prior(\vtheta) \propto \exp -\frac{1}{2\sigma^2} \vtheta^{\intercal}\vtheta,
\end{equation}
applying Bayes' rule gives:
\begin{equation}
    \log p(\vtheta \mid \dataone) = \log p(\dataone \mid \vtheta) + \log\prior(\vtheta) - \log p(\dataone).
\end{equation}
As $\log p(\dataone)$ is constant with respect to $\vtheta$, it does not influence the optimization problem. By replacing $\log p(\dataone \mid \vtheta)$ with the approximation from \cref{eq:taylor-approx} we obtain:\footnote{Note that $\log p(\dataone \mid \thetaone)$ in \cref{eq:taylor-approx} is also constant w.r.t. $\theta$, so we do not add it.}\looseness=-1\\
\begin{subequations} 
\allowdisplaybreaks
\setlength{\jot}{10pt}
    \begin{align}
    \log &p(\vtheta \mid \dataone) \approx \frac{1}{2}  (\vtheta - \thetaone)^\intercal \nabla^2_{\vtheta} \log p(\dataone \mid \thetaone)  (\vtheta - \thetaone) -\frac{1}{2\sigma^2} \vtheta^{\intercal}\vtheta \\
    \nonumber\\ 
    &= \frac{1}{2}  (\vtheta - \thetaone)^\intercal \left(\sum_{n=1}^N \left(\nabla^2_{\vtheta} \log p(\eos \mid \strone^{(n)}, \thetaone) + \sum_{t=1}^{T_n} \nabla^2_{\vtheta} \log p(x_{t}^{(n)} \mid \strone_{<t}^{(n)}, \thetaone)\right)\right)  (\vtheta - \thetaone) \nonumber  \\
   & \qquad \qquad \qquad \qquad\qquad \qquad\qquad\qquad\qquad \qquad\qquad\qquad \qquad\qquad\qquad \qquad-\frac{1}{2\sigma^2}
    \vtheta^{\intercal}\vtheta \\
    &\approx -\frac{1}{2}  (\vtheta - \thetaone)^\intercal \left( \sum_{n=1}^{N}\sum_{t=1}^{T_n+1}\E_{\overline{x} \sim p(\cdot \mid \strone_{<t}, \thetaone)}  \nabla^2_{\vtheta} -\log p(\overline{x} \mid \strone_{<t}^{(n)}, \thetaone) \right)  (\vtheta - \thetaone) -\frac{1}{2\sigma^2} \vtheta^{\intercal}\vtheta,     \label{eq:posterior-approx}
    \end{align}
\end{subequations}
where the last approximation replaces $\nabla^2_{\vtheta} \log p(x_{t}^{(n)} \mid \strone_{<t}^{(n)}, \thetaone)$ with its expectation $\E_{\overline{x} \sim p(\cdot \mid \strone_{<t}, \thetaone)}  \nabla^2_{\vtheta} \log p(\overline{x} \mid \strone_{<t}^{(n)}, \thetaone)$ and performs a sign manipulation.

\paragraph{A Fast Approximation.}
Exact computation of the matrix $\E_{\overline{x} \sim p(\cdot \mid \strone_{<t}, \thetaone)}  \nabla^2_{\vtheta} -\log p(\overline{x} \mid \strone_{<t}^{(n)}, \thetaone)$ is impractical for multiple reasons. 
First, it requires second-order automatic differentiation, which is relatively expensive and not a standard feature in common automatic differentiation toolkits, e.g., PyTorch \citep{pytorch}.
Second, recall that $p(\cdot \mid \strone_{<t}, \thetaone)$ is the next-symbol distribution of the language model, i.e., a distribution over $\overline{\Sigma}$, so $\E_{\overline{x} \sim p(\cdot \mid \strone_{<t}, \thetaone)}  \nabla^2_{\vtheta} -\log p(\overline{x} \mid \strone_{<t}^{(n)}, \thetaone)$ could be computed in $\mathcal{O}(|\overline{\Sigma}|)$ time.
While linear, $\overline{\Sigma}$ is often large in practice in modern language models. 
Thirdly, the matrix contains $\mathcal{O}(d^2)$ unique entries.\footnote{By Schwarz's lemma, if a function is twice \emph{continuously} differentiable, its Hessian is symmetric---hence, the big-$\mathcal{O}$.}
When $d$ is large, as it is in our case, we cannot compute all $d^2$ entries easily.
A classic identity involving the Fisher information matrix from \citet[pg. 185]{bickel2015mathematical} (see also \citet[Eq. 4]{NEURIPS2019_46a558d9}) allows us to derive the following approximation:
\newcommand{\gradient}{\mathbf{g}}
\begin{subequations} \label{eq:hessian-approx}
\begin{align}
\setlength{\jot}{10pt}
\sum_{n=1}^{N}\sum_{t=1}^{T_n+1}&\E_{\overline{x} \sim p(\cdot \mid \strone_{<t}, \thetaone)} \nabla^2_{\vtheta} -\log p(\overline{x} \mid \strone_{<t}^{(n)}, \thetaone)\nonumber\\
&\approx \sum_{n=1}^{N}\sum_{t=1}^{T_n+1}\E_{\overline{x} \sim p(\cdot \mid \strone_{<t}, \thetaone)} \nabla_{\vtheta} \log p(\overline{x} \mid \strone_{<t}^{(n)}, \thetaone) \nabla_{\vtheta} \log p(\overline{x} \mid \strone_{<t}^{(n)}, \thetaone)^{\intercal} \label{eq:step2} \\
&\approx \sum_{n=1}^{N}\sum_{t=1}^{T_n+1}\frac{1}{K} \sum_{k=1}^K \nabla_{\vtheta} \log p(\overline{x}^{(k)} \mid \strone_{<t}^{(n)}, \thetaone) \nabla_{\vtheta} \log p(\overline{x}^{(k)} \mid \strone_{<t}^{(n)}, \thetaone)^{\intercal} \label{eq:hessian-approx-monte-carlo}\\
&\defeq \widetilde{\fisher}_{\thetaone}
\end{align}
\end{subequations}
\Cref{eq:hessian-approx-monte-carlo} is a standard Monte Carlo approximation where $\overline{x}^{(k)} \sim p(\cdot \mid \strone_{<t}^{(n)}, \thetaone)$.
When $K \ll |\overline{\Sigma}|$, the sample-based approximation results in a significant speed-up. 
, to avoid $\bigO(d^2)$ computation time, we only approximate the \emph{diagonal} of $\widetilde{\fisher}_{\thetaone}$, which has  $\bigO(d)$ 
 entries.\looseness=-1

\paragraph{A Simple Regularizer.}
Synthesizing the above, we arrive at a simple regularizer that should promote a language model, previously trained on $\lone$ data, to retain its knowledge during training on $\ltwo$ data
\begin{equation}\label{eq:ewc-regularizer}
 \regularizer(\vtheta) =  \underbrace{\frac{1}{2} \sum_{i=1}^{d} \left(\widetilde{\fisher}_{\thetaone}\right)_{ii} \cdot (\vtheta - \vtheta^*_{\lone})_i^2}_{\regularizer_{\mathit{EWC}}} + \underbrace{\frac{1}{2\sigma^2}\vtheta^{\intercal}\vtheta}_{\regularizer_{\mathit{L2}}}
\end{equation}
The second term corresponds exactly to the well-known $L_2$ regularization term resulting from the prior over the parameters $\vtheta$. In practice, we generalize the coefficient $\frac{1}{2}$ into a tunable regularization coefficient $\lambda$ and the coefficient $\frac{1}{2\sigma^2}$ into a tunable regularization coefficient $\mu$.
We tune $\lambda$ on held-out data, but set $\mu = 0$ throughout the experiments and therefore omit it from the main text.\footnote{Based on empirical observations this hyperparameter did not have an effect on the results.}

\end{document}